\documentclass[lettersize,journal,biography]{IEEEtran}  
\usepackage{amsthm}
\usepackage{array}
\usepackage[caption=false,font=normalsize,labelfont=sf,textfont=sf]{subfig}
\usepackage{textcomp}
\usepackage{stfloats}
\usepackage{url}
\usepackage{verbatim}
\usepackage{cite}
\usepackage{multicol}
\usepackage{amsmath,amssymb,amsfonts}
\usepackage{xcolor}
\usepackage{mathrsfs}
\usepackage{lipsum} 
\usepackage{balance} 
\usepackage{cite}

\newtheorem{remark}{\bf Remark}[]

\usepackage{indentfirst}
\usepackage[ruled,vlined,linesnumbered]{algorithm2e}
\usepackage{flushend}
\usepackage{colortbl}
\usepackage{graphicx}  
\usepackage{booktabs}

\usepackage{multirow}
\usepackage{float}
\usepackage{nomencl}
\usepackage{etoolbox}
\usepackage{hyperref}

\newtheorem{lemma}{Lemma}
\usepackage{makecell}

\author{
\thanks{}
\thanks{}
}

\begin{document}

\title{ 
DepthGS: Depth-Guided Gaussian Splatting with Adaptive Regularization for High-Quality Surface Reconstruction }
\author{Xiao Ren, Yu Liu, Ning An, Jian Cheng, Xin Qiao, You-Fu Li, \IEEEmembership{Fellow,~IEEE,} and He Kong
\thanks{This work was supported by National Key R\&D
Program of China under Grant No. 2024YFB4710900, the National
Natural Science Foundation of China under Grant
No. U24A20265, the Shenzhen Science and Technology Program under Grant KQTD20221101093557010, and the Guangdong Science and Technology Program under Grant 2024B1212010002. Corresponding author: He Kong.}
\thanks{Xiao Ren, Yu Liu, and He Kong are with 
School of Automation and Intelligent Manufacturing, Southern University of Science and Technology, Shenzhen 518055, China; Emails: [12431359,12250044]@mail.sustech.edu.cn; kongh@sustech.edu.cn. Ning An and Jian Cheng are with the Research Institute of Mine Artificial Intelligence, China Coal Research Institute, and the State Key Laboratory of Intelligent Coal Mining and Strata Control, Beijing 100031, China; Emails: ning.an.010@foxmail.com; jiancheng@tsinghua.org.cn. Xin Qiao is with the Institute of Artificial Intelligence and Robotics, Xi’an Jiaotong University, Xi’an 710049, China; Email: wudiqx@xjtu.edu.cn. You-Fu Li is with the Department of Mechanical Engineering, City University of Hong Kong, Hong Kong SAR, China. Email: meyfi@cityu.edu.hk.}}

\maketitle
\thispagestyle{empty}
\pagestyle{empty}

\begin{abstract}
Recently, 3D Gaussian Splatting (3DGS) has emerged as an efficient
representation for novel view synthesis, offering fast differentiable
rendering and high visual fidelity, and has increasingly been extended
to surface reconstruction. 
However, the relationship between depth information and reconstruction error remains insufficiently explored,  limiting the principled design of geometric optimization strategies.
To fill the above gap, we propose DepthGS, a depth-guided Gaussian splatting framework. 
We derive theoretical relations between depth discrepancies and reconstruction error under single-view and multi-view settings, and rigorously establish corresponding upper bounds on reconstruction error. The above analysis informs the design of adaptive regularization strategies in both single-view and multi-view optimization. More specifically, for single-view optimization, we propose a depth-difference-guided adaptive normal regularization strategy that modulates the normal constraint strength using a geometric reliability weight computed from local depth differences. For multi-view optimization, we introduce a cross-view depth consistency regularization strategy that promotes consistency between neighboring-view depth maps and the corresponding depth predictions obtained by transforming the current view’s depth predictions into those neighboring views.
Extensive experiments on DTU, Mip-NeRF360, and Tanks and Temples demonstrate that DepthGS achieves competitive performance in both surface reconstruction and novel view synthesis.

\end{abstract}


\section{INTRODUCTION}

Multi-view 3D reconstruction plays a crucial role in computer vision, with applications ranging from robotics~\cite{GARAD-SLAM,Dy3DGS-SLAM,wakulicz2021active} and autonomous driving~\cite{eiffert2022resource,wang2024slam,AutoSplat} to augmented and virtual reality~\cite{lianggus_ir,wangmba_slam,yin_ms_nerf}.
Neural radiance fields (NeRFs)~\cite{mildenhall2021nerf,TD-NeRF,Super-NeRF} represent scenes using continuous radiance fields and achieve high-quality novel view synthesis and geometric reconstruction. 
However, their computationally expensive volume rendering limits real-time applications. 
In contrast, 3D Gaussian Splatting (3DGS)~\cite{kerbl20233d} represents scenes using explicit anisotropic Gaussians and enables efficient optimization and real-time high-fidelity rendering through differentiable $\alpha$-blending.

Despite its advances in rendering speed and quality, 3DGS still faces challenges in accurate surface reconstruction. The inherent irregularity of Gaussian primitives makes it difficult to precisely fit real-world surfaces, which may lead to geometric artifacts~\cite{chengtopo4D}.


Geometry-aware approaches~\cite{li2024geogaussian,cheng2024gaussianpro}
improve Gaussian representations through additional geometric constraints, while recent methods incorporating depth and normal cues
~\cite{turkulainen2024dn,chen2024pgsr,chen2025vcr} demonstrate the effectiveness of such geometric information for surface reconstruction. Nevertheless, the reconstruction errors of the three-dimensional points obtained from unbiased depth and rendered distance through
inverse projection have not been explicitly characterized with respect to the corresponding ground-truth surface points.
Moreover, the cross-view consistency of these two depth representations remains underexplored. These limitations motivate us to derive reconstruction-error bounds and use the resulting depth discrepancies to guide adaptive geometric regularization.

In this paper, we propose DepthGS, a depth-guided Gaussian splatting framework with adaptive regularization for high-quality surface reconstruction. Through mathematical analysis, we investigate the relationship between depth information and reconstruction error and derive the corresponding upper bounds. Based on the single-view analysis, we propose a depth-difference-guided adaptive normal regularization strategy. 
For multi-view scenes, we propose a cross-view depth consistency regularization strategy to constrain depth 
discrepancies between neighboring views and improve multi-view geometric consistency. Our main contributions are summarized as follows:
\begin{itemize}
\item Through mathematical analysis, we derive the relationships between
depth discrepancies and reconstruction error under single-view and multi-view conditions, together with the upper bounds for reconstruction error, providing theoretical guidance for the subsequent optimization strategies.
\item Based on the single-view error analysis, we propose a depth-difference-guided adaptive normal regularization strategy that modulates the normal constraint strength according to the local depth difference. Specifically, weaker constraints are applied to regions with larger depth differences, whereas stronger constraints are applied to regions with smaller depth differences.
\item Based on the multi-view error analysis, we propose a cross-view depth consistency regularization strategy that reduces depth discrepancies between neighboring views, thereby improving multi-view geometric consistency.
\item Extensive experiments on Mip-NeRF360~\cite{barron2022mip}, DTU~\cite{jensen2014large} and Tanks and Temples~\cite{knapitsch2017tanks} demonstrate that DepthGS achieves competitive performance in both novel view synthesis and surface reconstruction.
\end{itemize}

\section{RELATED WORK}

\subsection{ Novel View Synthesis}

Novel view synthesis aims to render unseen views while preserving scene structure and visual details. 
Classical Structure-from-Motion and Multi-View Stereo methods recover scene geometry through multi-view correspondences and epipolar constraints~\cite{schonberger2016structure}. 
Soft3D~\cite{penner2017soft} introduces a differentiable density-field representation for view synthesis, while NeRF and its variants represent scenes using continuous radiance fields and achieve high-quality rendering~\cite{deng_fov_nerf,huang_ltm_nerf}. 
However, their volumetric formulations generally incur considerable computational costs.

3DGS provides an efficient alternative by representing scenes using explicit anisotropic Gaussians and differentiable rasterization~\cite{efficient_geometric,GeoRGS}. 
Subsequent approaches improve representation compactness and rendering efficiency~\cite{Compact_3D_Gaussian,Steepest_Descent,mallick2024taming}, enhance structural integrity and large-scale rendering~\cite{shen2024topology,feng2025flashgs}, and extend Gaussian representations to dynamic scenes~\cite{Motion-Aware,Fov-GS}.
Nevertheless, improvements in rendering quality and efficiency do not necessarily ensure accurate and view-consistent surface geometry.

\begin{figure*}[t]
    \centering
    \includegraphics[width=0.95\textwidth]{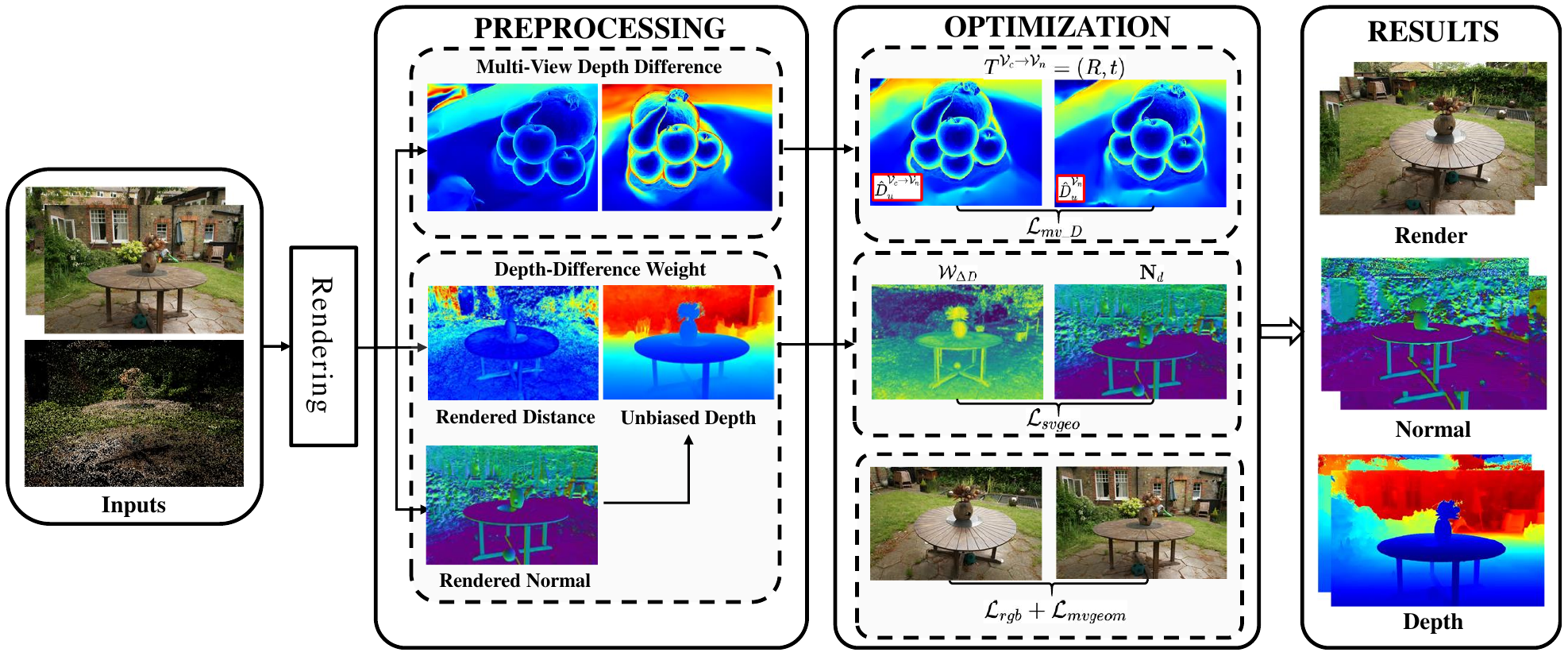}
    \caption{Overview of the proposed framework. The Gaussian representation is initialized from a sparse point cloud and used to render the distance and normal maps. Unbiased depth maps are then derived from these rendered quantities. The discrepancy between the unbiased depth and rendered distance is used to compute the depth-difference-based weight $\mathcal{W}_{\Delta D}$, which modulates the single-view normal constraint
    $\mathcal L_{svgeo}$. In the multi-view branch, coordinate transformations establish depth correspondences between neighboring views to construct the cross-view depth consistency constraint $\mathcal L_{mv\_D}$. The proposed constraints are jointly optimized with the baseline objectives to improve surface  reconstruction and multi-view geometric consistency.}
    \label{fig:Framework}
\end{figure*}

\subsection{Gaussian Splatting Surface Reconstruction}

With the increasing use of Gaussian splatting in novel view synthesis, its potential for surface reconstruction has also been explored. 2DGS~\cite{huang20242d} replaces volumetric Gaussian ellipsoids with planar Gaussian disks to achieve more consistent geometric representations, while GOF~\cite{yu2024gaussian} extracts surfaces from Gaussian opacity fields.

Recent approaches further improve reconstruction through geometry-oriented representations and surface constraints.
GFSGS~\cite{jiang2025geometry} and GSDF~\cite{yu2024gsdf} introduce geometry-aware fields and distance representations to regularize Gaussian optimization. GS-Pull~\cite{zhang2024gspull} and GS-UDF~\cite{li2025gaussianudf} infer signed or unsigned distance functions from Gaussian primitives. QGS~\cite{zhang2025quadratic}
models surfaces using second-order geometric primitives, while GSPrior~\cite{noda20263d} incorporates self-constrained geometric priors for high-fidelity surface reconstruction. 

Methods incorporating depth and normal cues provide complementary geometric information. 
DN-Splatter~\cite{turkulainen2024dn} incorporates depth and normal priors for Gaussian splatting and meshing. 
VCR-GauS~\cite{chen2025vcr} couples depth and normal information through a view-consistent regularizer. 
PGSR~\cite{chen2024pgsr} introduces unbiased-depth rendering and
multi-view regularization for efficient and high-fidelity surface reconstruction.

These studies demonstrate the effectiveness of depth and normal cues for geometric regularization. 
However, existing Gaussian splatting methods incorporating depth and normal cues do not explicitly derive reconstruction-error bounds that involve the discrepancy between
unbiased depth and rendered distance. 
Moreover, the multi-view consistency of these two depth representations across neighboring views remains underexplored. 
DepthGS addresses these limitations by deriving the relationships between depth discrepancies and reconstruction errors under single-view and multi-view conditions. 
Based on this analysis, we introduce a depth-difference-guided adaptive normal regularization strategy and a cross-view depth consistency regularization strategy.

\section{Preliminaries on 3D Gaussian Splatting}

Following 3DGS~\cite{kerbl20233d}, a scene is represented by a set of anisotropic Gaussian primitives. The $i$-th Gaussian is defined by its center $\mu_i$ and covariance matrix
$\Sigma_i$:
\begin{equation}
    G(\mu;\mu_i,\Sigma_i) = \exp\left[-\frac{1}{2} (\mu-\mu_i)^\top \Sigma_i^{-1}
    (\mu-\mu_i) \right].
    \label{eq:3d_gaussian}
\end{equation}
To ensure positive semidefiniteness, the covariance is parameterized as 
$\Sigma_i=R_iS_iS_i^\top R_i^\top$, where $R_i$ and $S_i$ denote the rotation and scaling matrices, respectively. Under a given camera view, the Gaussians are projected onto the image plane and composited in depth order through differentiable $\alpha$-blending.

At pixel $p$, the rendered color and normal are computed as
\begin{equation}
    \begin{aligned}
        \widehat{C}(p) = \sum_{i\in N} T_i\alpha_i c_i, \quad
        \mathbf N(p) = \sum_{i\in N} T_i\alpha_i R_c^\top n_i,
    \end{aligned}
    \label{eq:color_normal_rendering}
\end{equation}
where $T_i=\prod_{j=1}^{i-1}(1-\alpha_j)$ denotes the accumulated transmittance, $\alpha_i$ is the contribution weight of the $i$-th Gaussian, $c_i$ denotes its view-dependent appearance, and $N$ is the ordered set of Gaussians contributing to pixel $p$. 
Moreover, $n_i$ denotes the Gaussian normal, and $R_c^\top n_i$ denotes the corresponding normal expressed in the camera coordinate system.

Following PGSR~\cite{chen2024pgsr}, the renderer further produces the rendered distance and unbiased depth:
\begin{equation}
    \begin{aligned}
        D_{\mathcal{D}}(p) = \sum_{i\in N}T_i\alpha_i d_i, \quad 
        D_u(p)  =  \frac{D_{\mathcal{D}}(p)} {\mathbf{N}(p)\cdot v(p)}.
    \end{aligned}
    \label{eq:mathcal_D}
\end{equation}
Here, $d_i=(R_c^\top n_i)^\top R_c^\top(\mu_i-T_c)$ denotes the plane-to-camera distance of the $i$-th Gaussian, and $T_c$ is the camera center. For pixel $p=(u,v)$ with homogeneous coordinate $\tilde{\mathbf{p}}=[u,v,1]^\top$,
$v(p)=K^{-1}\tilde{\mathbf{p}}/ \|K^{-1}\tilde{\mathbf{p}}\|_2$ is the corresponding unit camera-ray direction, where $K$ denotes the camera intrinsic matrix.

\section{METHODOLOGY}
The framework proposed in this paper comprises two major components, which address geometric constraint optimization under single-view and multi-view conditions, respectively. The overall framework of our algorithm is depicted in Fig.~\ref{fig:Framework}. 

\subsection{ Single-View Geometric Constraint Optimization }
\label{subsec:single_view_optimization}

\subsubsection{Single-View Depth Difference-Reconstruction Error Relationship}
\begin{figure}[!t]
    \centering
    \includegraphics[width=\columnwidth]
    {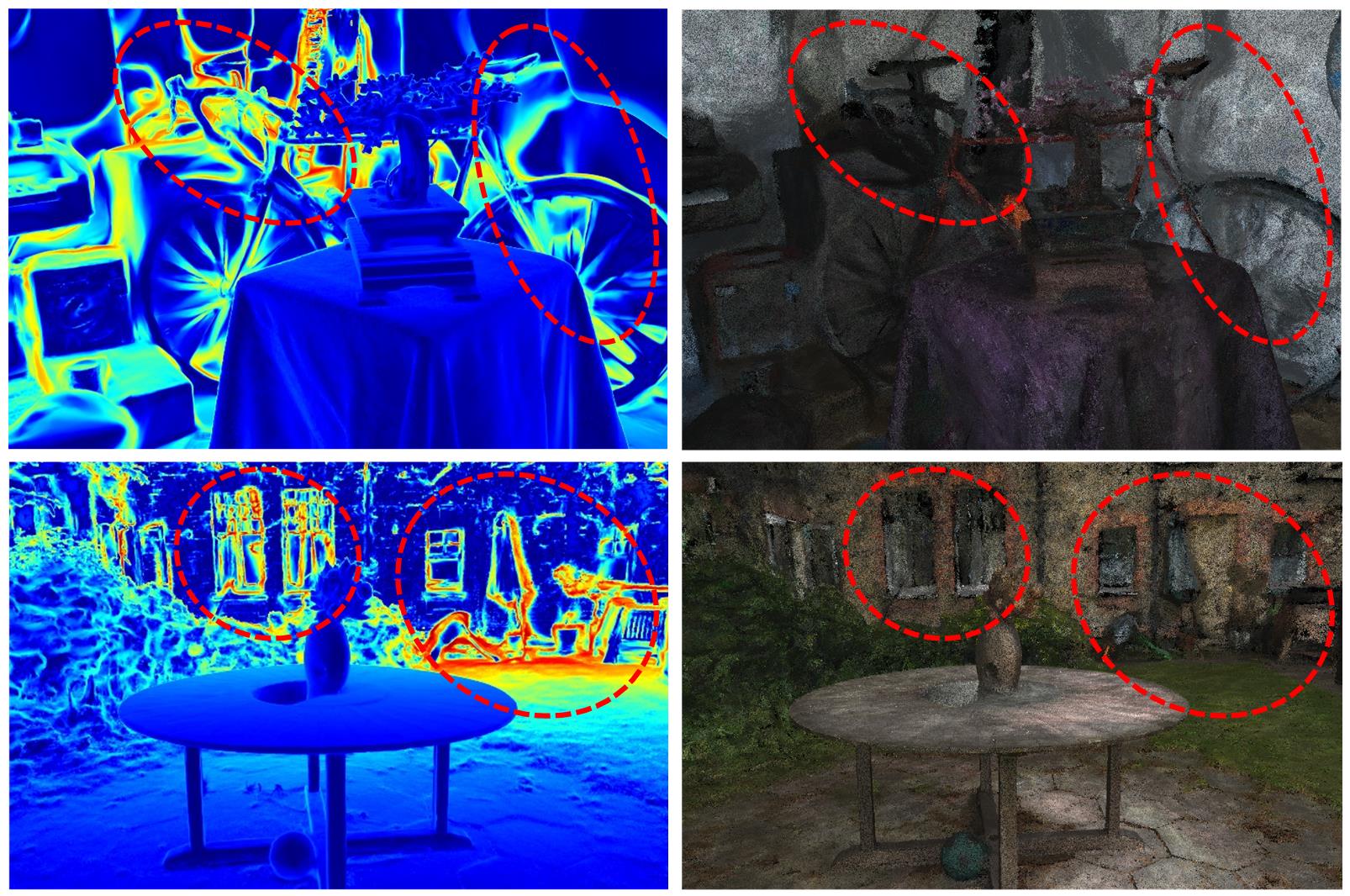}
    \caption{Depth differences (left) and corresponding Gaussian distributions (right); large discrepancies often coincide with irregular Gaussian layouts.}
    \label{fig:depth_diff_ellpisods}
\end{figure}

As shown in Eq.~\ref{eq:mathcal_D}, Gaussian splatting produces an unbiased depth $D_u$ and a rendered distance $D_{\mathcal D}$.
The corresponding three-dimensional points are obtained from $D_u$ and $D_{\mathcal D}$ through inverse projection along the same camera ray. Their Euclidean distance reflects the discrepancy between the two depth representations. 
As illustrated in Fig.~\ref{fig:depth_diff_ellpisods}, large depth differences are often
observed in regions with irregular Gaussian distributions.

To formalize this observation, we analyze the reconstruction errors of the corresponding three-dimensional points obtained from $D_u$ and $D_{\mathcal D}$ with respect to the ground-truth surface point. We first derive the single-view relationship between the depth  difference and reconstruction error and then extend the analysis to the multi-view case.

\begin{lemma}
{\rm 
In a single-view scene, the aggregate reconstruction error associated with the unbiased depth satisfies
\begin{equation}
    \begin{aligned}
        E_{sv} &\triangleq \int_{\Omega} w(p) \left(\epsilon_u^{sv}(p) \right)^2
        dp \\
        &\leq \int_{\Omega} w(p) \left(\epsilon_{\mathcal{D}}^{sv}(p) +      |\Delta D(p)| \right)^2 dp,
    \end{aligned}
\label{eq:lemma1}
\end{equation}
where $w(p)\geq0$, $\Omega$ denotes the valid image domain, and $\epsilon_u^{sv}(p)$ and $\epsilon_{\mathcal{D}}^{sv}(p)$ denote the reconstruction errors associated with the unbiased depth and
rendered distance, respectively.
}
\end{lemma}

\begin{proof}
{\rm 
The depth difference is given by
\begin{equation}
    \Delta D(p)
    =
    D_u(p)-D_{\mathcal{D}}(p)
    =
    \frac{D_{\mathcal{D}}(p)}
    {\mathbf{N}(p)\cdot v(p)}
    -
    D_{\mathcal{D}}(p).
    \label{eq:D_p}
\end{equation}
The corresponding three-dimensional points obtained through inverse projection are denoted by $X_u$ and $X_{\mathcal D}$, respectively:
\begin{align}
    X_u &= \pi^{-1}\!\left(p,D_u(p)\right) = \frac{D_{\mathcal D}(p)} {\mathbf N(p)\cdot v(p)}v(p),
    \label{eq:X_u}
    \\
    X_{\mathcal D} &= \pi^{-1}\!\left(p,D_{\mathcal D}(p)\right) = D_{\mathcal D}(p)v(p).
    \label{eq:X_DD}
\end{align}
Here, $v(p)$ is the unit camera-ray direction defined in Eq.~\ref{eq:mathcal_D}. Let $X_{\mathrm{true}}$ denote the corresponding ground-truth surface point. The reconstruction errors associated with the unbiased depth and rendered distance are defined as
\begin{align}
    \epsilon_u^{sv}(p) &= \left\|X_u-X_{\mathrm{true}}\right\|_2 =
    \left\| \pi^{-1}\!\left(p,D_u(p)\right) - X_{\mathrm{true}} \right\|_2,
    \\
    \epsilon_{\mathcal D}^{sv}(p) &= \left\|X_{\mathcal D}-X_{\mathrm{true}}\right\|_2
    = \left\| \pi^{-1}\!\left(p,D_{\mathcal D}(p)\right) - X_{\mathrm{true}} \right\|_2.
\end{align}

Application of the triangle inequality to $X_u-X_{\mathrm{true}}=(X_u-X_{\mathcal D})+(X_{\mathcal D}-X_{\mathrm{true}})$, together with the above error definitions, yields
\begin{equation}
    \epsilon_u^{sv}(p) \leq  \epsilon_{\mathcal D}^{sv}(p) + \left\|
        \pi^{-1}\!\left(p,D_u(p)\right) -  \pi^{-1}\!\left(p,D_{\mathcal D}(p)\right)
    \right\|_2.
    \label{eq:epsilon_u}
\end{equation}
Using Eqs.~\ref{eq:X_u}--\ref{eq:X_DD} and $\|v(p)\|_2=1$, the second term on the right-hand side of Eq.~\ref{eq:epsilon_u} can be expressed as
\begin{equation}
    \begin{aligned}
        &
        \left\| \pi^{-1}\!\left(p,D_u(p)\right) - \pi^{-1}\!\left(p,D_{\mathcal D}(p)\right)
        \right\|_2
        \\
        &\qquad
        = \left|D_{\mathcal D}(p)\right| \left| \frac{1}{\mathbf N(p)\cdot v(p)}-1 \right|
        = |\Delta D(p)|.
    \end{aligned}
    \label{eq:Np_vp}
\end{equation}

Substituting Eq.~\ref{eq:Np_vp} into
Eq.~\ref{eq:epsilon_u} yields
\begin{equation}
    \epsilon_u^{sv}(p)
    \leq
    \epsilon_{\mathcal D}^{sv}(p)
    +
    |\Delta D(p)|.
    \label{eq:epsilon_u_sv}
\end{equation}
Since both sides of Eq.~\ref{eq:epsilon_u_sv} are nonnegative, squaring the inequality, multiplying both sides by the nonnegative weight $w(p)$, and integrating over the valid image domain $\Omega$ yield the aggregate reconstruction-error bound in Eq.~\ref{eq:lemma1}. This completes the proof.
}
\end{proof}

\begin{remark}
Expanding the right-hand side of Eq.~\ref{eq:lemma1} gives
\begin{equation}
    \begin{aligned}
    E_{sv}\leq \int_{\Omega} w(p)\Big[ &\left(\epsilon_{\mathcal D}^{sv}(p)\right)^2
    +2\left|\epsilon_{\mathcal D}^{sv}(p)\right| \left|\Delta D(p)\right| \\
    &+\left|\Delta D(p)\right|^2 \Big]\,dp.
    \end{aligned}
    \label{eq:expanding}
\end{equation}
Thus, for a fixed $\epsilon_{\mathcal D}^{sv}(p)$, reducing $|\Delta D(p)|$ tightens the derived reconstruction-error bound, motivating the adaptive regularization below.
\end{remark}

\subsubsection{ Depth-Difference-Weighted Adaptive Normal Regularization}

Eqs.~\ref{eq:Np_vp} and~\ref{eq:epsilon_u_sv} show that $|\Delta D(p)|$ equals the Euclidean distance between $X_u$ and $X_{\mathcal D}$ and appears explicitly in the derived
reconstruction-error bound. Thus, for a fixed or comparable $\epsilon_{\mathcal D}^{sv}(p)$, a larger depth difference enlarges the right-hand side of the bound and indicates a larger discrepancy between the unbiased depth $D_u(p)$ and the rendered distance $D_{\mathcal D}(p)$.

Guided by this relationship, we use $|\Delta D(p)|$ to adaptively weight the normal constraint between the reference normal $\mathbf N_d(p)$ computed from $D_u(p)$ and the rendered normal $\mathbf N(p)$.

\begin{figure}[h!]
    \centering
    \includegraphics[width=\linewidth]{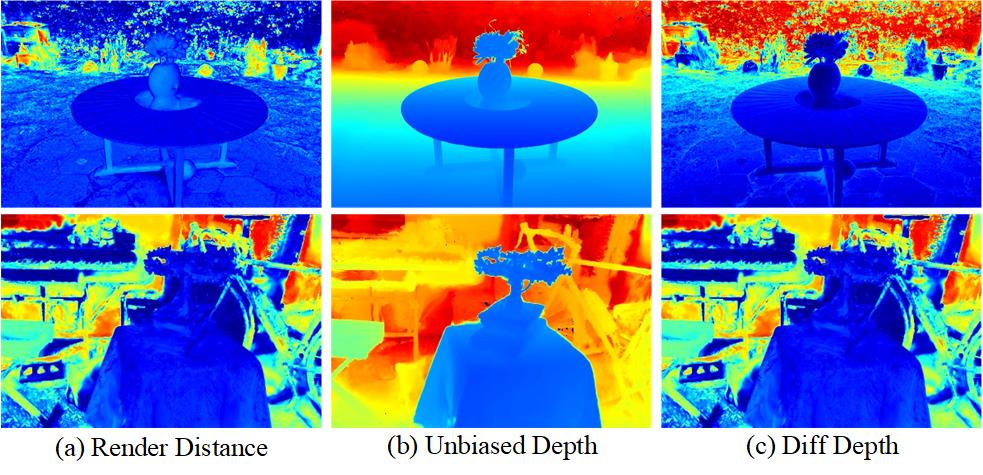}
    \caption{Depth-difference weights are derived from the discrepancy between the rendered-distance map and the unbiased-depth map. The greater the depth difference, the smaller the weight.}
    \label{fig:depth_diff}
\end{figure}

As illustrated in Fig.~\ref{fig:depth_diff}, rather than directly minimizing $|\Delta D(p)|$, we treat it as a local geometric reliability measure and map it to the adaptive weight
$\mathcal W_{\Delta D}(p)$. The weight is defined as
\begin{equation}
    \mathcal{W}_{\Delta D}(p) = 1 - \frac{|\Delta D(p)| - \Delta D_{\min}}{ \Delta D_{\max} - \Delta D_{\min} + \sigma},
    \label{eq:delta_W(p)}
\end{equation}
where $\sigma>0$ prevents division by zero, and $\Delta D_{\min}$/$\Delta D_{\max}$ are the minimum/maximum of $|\Delta D(p)|$ over valid pixels. A smaller depth difference produces a larger weight and hence stronger normal regularization, whereas a larger difference weakens the constraint to avoid propagating unreliable local geometry.

\begin{figure}[t!]
    \centering
    \includegraphics[width=0.5\textwidth]{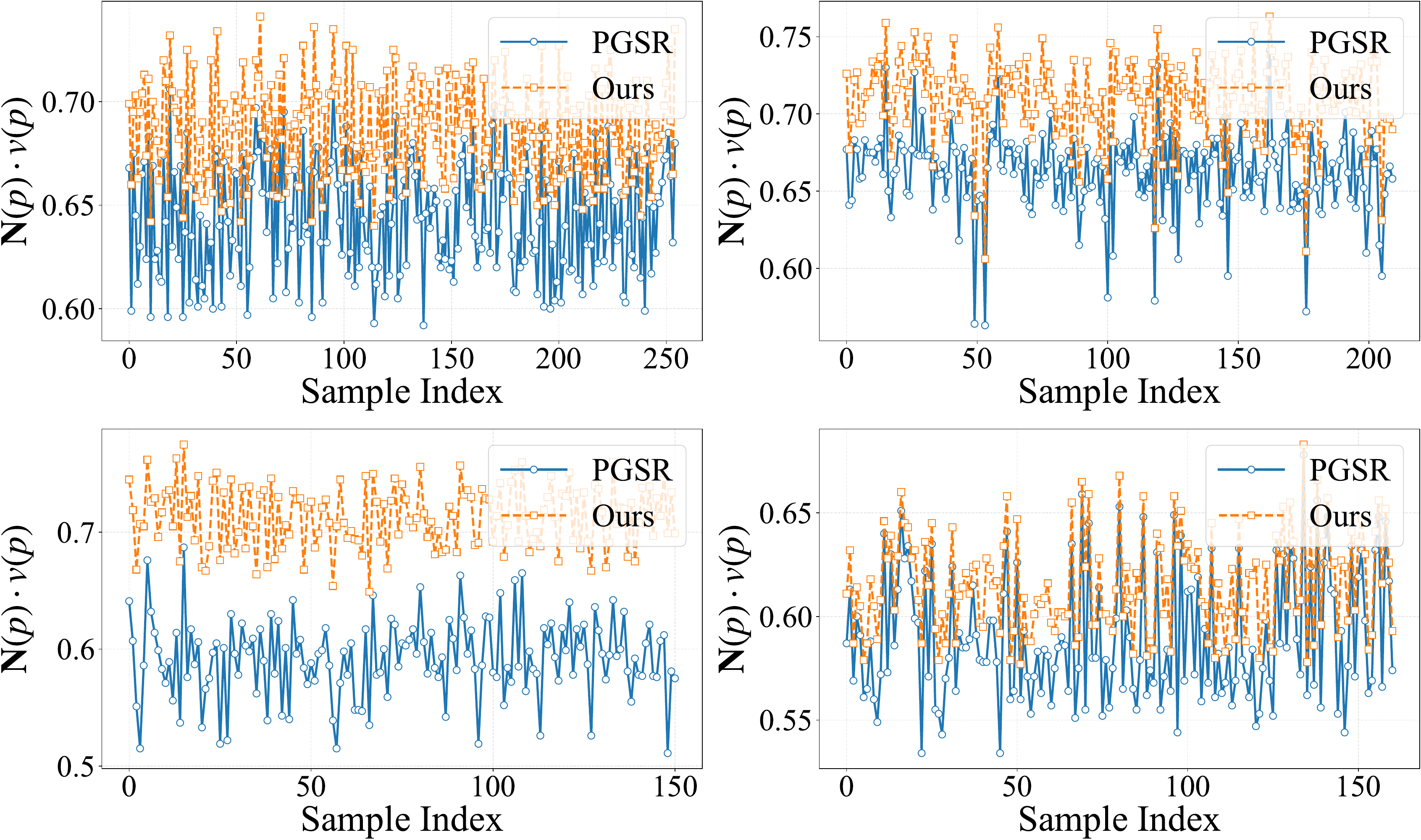}
    \caption{Comparison of $\mathbf{N}(p)\cdot v(p)$ obtained by PGSR         and the proposed method. The proposed method generally yields              normal--ray consistency values closer to 1 in the evaluated scenes.}
    \label{fig:camera_ray_cos_curve}
\end{figure}

The single-view normal constraint is defined as
\begin{equation}
    \mathcal{L}_{svgeo} = \frac{1}{|\Omega_{D_{\mathcal{D}}}|} \sum_{p \in \Omega_{D_{\mathcal{D}}}} \mathcal{W}(p)
    \left\| \mathbf{N}_d(p)-\mathbf{N}(p) \right\|_1,
    \label{eq:L_svgeo}
\end{equation}
where $\Omega_{D_{\mathcal{D}}}$ denotes the valid pixel domain of the
rendered-distance map, $\mathbf{N}_d$ represents the reference normals
generated from unbiased depth, and $\mathbf{N}(p)$ denotes the
rendered normal.
The complete weight is $\mathcal W(p)= \mathcal W_{\Delta D}(p) \left(\overline{\nabla I}(p)\right)^2$, where $\mathcal W_{\Delta D}$ is the proposed depth-difference-based reliability weight, while $\left(\overline{\nabla I}(p)\right)^2$ is the edge-aware modulation inherited from PGSR~\cite{chen2024pgsr}. 
As shown in Fig.~\ref{fig:camera_ray_cos_curve}, the normal–ray consistency values obtained by PGSR~\cite{chen2024pgsr} and the proposed method are compared across selected scenes.

\subsection{Multi-View Geometric Constraint Optimization}
\label{subsec:multi_view_optimization}

\subsubsection{Multi-View Depth Difference-Reconstruction Error Relationship}

Corresponding three-dimensional points observed from different views should satisfy rigid-transformation consistency. However, as shown in Fig.~\ref{fig:multi_view_depth}, neighboring views may exhibit different depth discrepancies at corresponding surface locations. This motivates us to relate cross-view depth-difference residuals to multi-view reconstruction error.

\begin{lemma}
{\rm
In multi-view scenes, the aggregate reconstruction error associated with the unbiased-depth-based point clouds satisfies
\begin{equation}
    \begin{aligned}
        E_{mv} &\triangleq \sum_{(i,j)\in\mathcal P} \int_{\Omega_{ij}} w_{ij}(p) \left(
        \epsilon_{ij}^{mv}(p) \right)^2 dp \\
        & \leq 2 \sum_{(i,j)\in\mathcal P} \int_{\Omega_{ij}} w_{ij}(p) \left[ \left(
        \epsilon_{\mathcal{D},ij}^{mv}(p) \right)^2 + r_{\Delta,ij}^2(p) \right]
        dp,
    \end{aligned}
    \label{eq:multi_view_bound}
\end{equation}
where $\mathcal P$ denotes the selected directed current/nearest-neighbor view pairs, $R_{ij}$ is the rotation from view $i$ to view $j$, $w_{ij}(p)\geq0$, and $\Omega_{ij}$ is the
valid correspondence domain. Moreover, $\epsilon_{ij}^{mv}(p)$ and $\epsilon_{\mathcal D,ij}^{mv}(p)$ are the unbiased- and rendered-distance-based cross-view discrepancies, 
respectively, and $r_{\Delta,ij}(p)=\left\|\Delta D^{\mathcal V_j}(p)v^{\mathcal V_j}(p)-
R_{ij}\!\left(\Delta D^{\mathcal V_i}(p)v^{\mathcal V_i}(p)\right)\right\|_2$ is the corresponding depth-difference residual. 
}
\end{lemma}
\begin{proof}
{\rm
Let the rigid transformation from the current view $\mathcal V_c$ to its nearest-neighbor view $\mathcal V_n$ be $T_{\mathcal V_c}^{\mathcal V_n}=(R,t)$, where $R\in\mathrm{SO}(3)$ and $t\in\mathbb R^3$.
Let $p_c=(u_c,v_c)$ and $p_n=(u_n,v_n)$ denote corresponding pixels in the two views, with unit camera-ray directions $v^{\mathcal V_c}$ and $v^{\mathcal V_n}$, respectively.
For clarity, the view-pair and pixel indices are omitted in the following derivation.
\begin{figure}[t!]
    \centering
    \includegraphics[width=\linewidth]{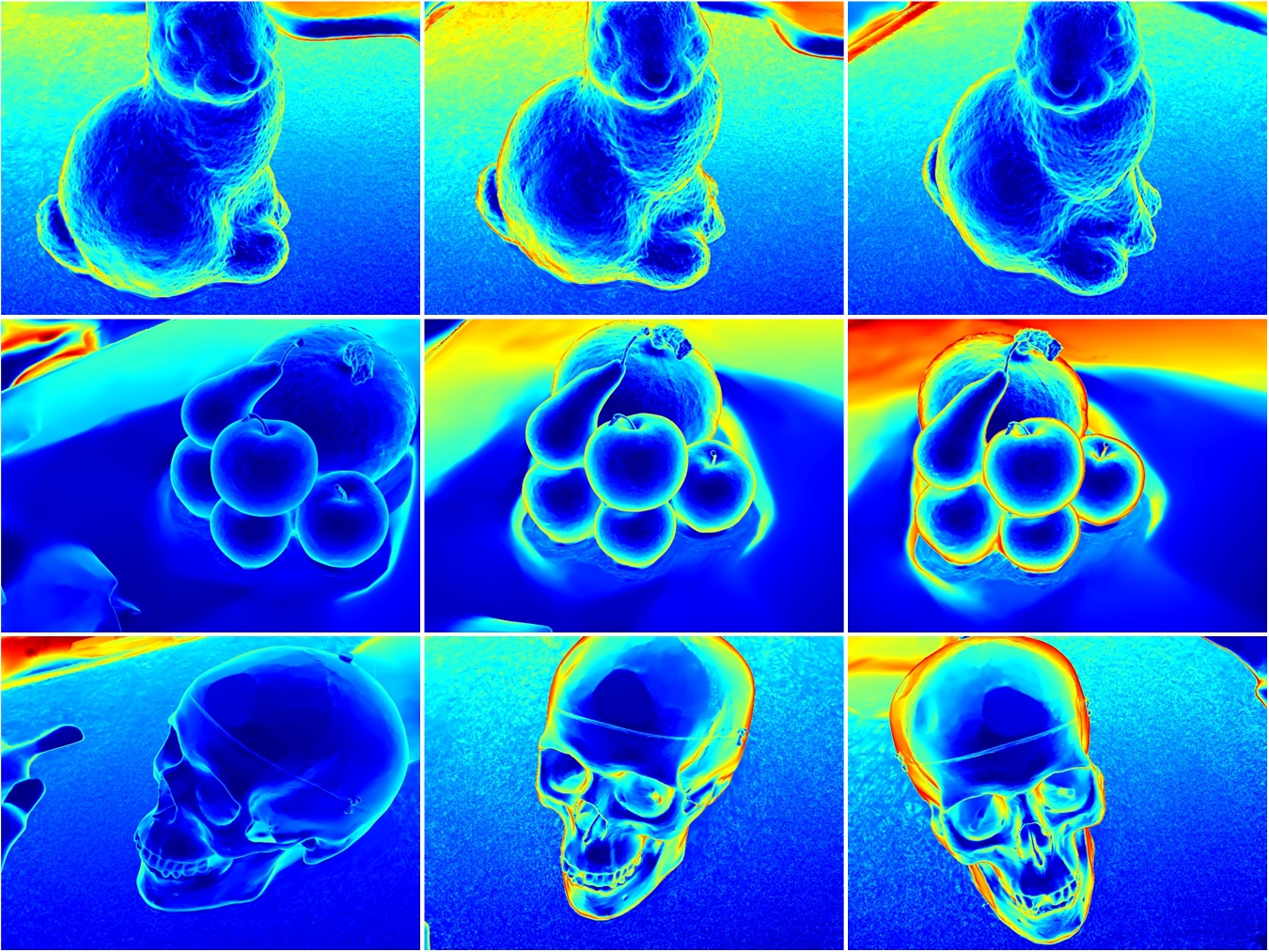}
    \caption{Cross-view inconsistency of depth-difference maps at corresponding surface locations in neighboring views.}
    \label{fig:multi_view_depth}
\end{figure}
Consistent with the single-view definition, the three-dimensional points obtained from the unbiased depth and rendered distance are
\begin{equation}
\begin{aligned}
    X_u^{\mathcal{V}_c} &=  D_u^{\mathcal{V}_c}v^{\mathcal{V}_c},
    &
    X_{\mathcal{D}}^{\mathcal{V}_c} &= D_{\mathcal{D}}^{\mathcal{V}_c}v^{\mathcal{V}_c},
    \\
    X_u^{\mathcal{V}_n} &=
    D_u^{\mathcal{V}_n}v^{\mathcal{V}_n},
    &
    X_{\mathcal{D}}^{\mathcal{V}_n} &=
    D_{\mathcal{D}}^{\mathcal{V}_n}v^{\mathcal{V}_n}.
\end{aligned}
\end{equation}
Here, $D_{\mathcal{D}}^{\mathcal{V}_c}$ and $D_{\mathcal{D}}^{\mathcal{V}_n}$ represent the rendered distance maps under the current view and the nearest-neighbor view, respectively.
To relate the exact vector residual $r_\Delta$ to a scalar cross-view depth-difference residual, we adopt a shared-target-ray construction, under which the transformed unbiased-depth and rendered-distance representations are projected onto the target ray $v^{\mathcal V_n}$. The corresponding projected depths are
\begin{align}
    \hat{D}_{u}^{\mathcal{V}_n} = {v^{\mathcal{V}_n}}^\top (R D_{u}^{\mathcal{V}_c} v^{\mathcal{V}_c} + t) , \label{eq:D_v_n_u} \\
    \hat{D}_{\mathcal{D}}^{\mathcal{V}_n} = {v^{\mathcal{V}_n}}^\top (R {D_{\mathcal{D}}^{\mathcal{V}_c}} v^{\mathcal{V}_c} + t) . \label{eq:D_v_n_d} 
\end{align}
To quantify multi-view consistency, the Euclidean distance between corresponding points after coordinate transformation is calculated.
Accordingly, the cross-view reconstruction error associated with the unbiased-depth-based points is defined as
\begin{equation}
    \begin{aligned}
        \epsilon^{mv} &=\left\|X_u^{\mathcal{V}_n} - X_u^{\mathcal{V}_c\rightarrow\mathcal{V}_n} \right\|_2
        \\
        &= \left\|D_u^{\mathcal{V}_n}v^{\mathcal{V}_n} - \left(R D_u^{\mathcal{V}_c}v^{\mathcal{V}_c} + t \right) \right\|_2.
    \end{aligned}
    \label{eq:epsilon_mv}
\end{equation}
Define the depth differences as $\Delta D^{\mathcal{V}_c} = D_u^{\mathcal{V}_c} -
D_{\mathcal{D}}^{\mathcal{V}_c}$ and $\Delta D^{\mathcal{V}_n} = D_u^{\mathcal{V}_n} - D_{\mathcal{D}}^{\mathcal{V}_n}$.
Accordingly, the unbiased depths can be expressed as $D_u^{\mathcal{V}_c} = D_{\mathcal{D}}^{\mathcal{V}_c} + \Delta D^{\mathcal{V}_c}$ and $D_u^{\mathcal{V}_n} = D_{\mathcal{D}}^{\mathcal{V}_n}
+ \Delta D^{\mathcal{V}_n}$.

Substituting these expressions into Eq.~\ref{eq:epsilon_mv} yields:
\begin{equation}
\begin{aligned}
    \epsilon^{mv} = \Bigl\| &D_{\mathcal{D}}^{\mathcal{V}_n}     v^{\mathcal{V}_n} - R D_{\mathcal{D}}^{\mathcal{V}_c}        v^{\mathcal{V}_c} - t
        \\
        &+
        \left( \Delta D^{\mathcal{V}_n} v^{\mathcal{V}_n} - R\Delta D^{\mathcal{V}_c} v^{\mathcal{V}_c} \right) \Bigr\|_2 .
\end{aligned}
\end{equation}
Applying the triangle inequality yields:
\begin{equation}
    \epsilon^{mv}
    \leq
    \epsilon_{\mathcal{D}}^{mv}
    +
    \left\|
    \Delta D^{\mathcal{V}_n}v^{\mathcal{V}_n}
    -
    R\Delta D^{\mathcal{V}_c}v^{\mathcal{V}_c}
    \right\|_2,
    \label{eq:epsilon_mv_2}
\end{equation}
where $\epsilon_{\mathcal{D}}^{mv} \triangleq \left\| 
X_{\mathcal{D}}^{\mathcal{V}_n} - X_{\mathcal{D}}^{\mathcal{V}_c\to\mathcal{V}_n}
\right\|_2 = \left\| D_{\mathcal{D}}^{\mathcal{V}_n}v^{\mathcal{V}_n} -
R\!\left(D_{\mathcal{D}}^{\mathcal{V}_c}v^{\mathcal{V}_c}\right) - t \right\|_2$
denotes the Euclidean distance between the rendered-distance-based point in the neighboring view and the corresponding current-view point transformed into the neighboring-view coordinate system.
For brevity, we denote the second term on the right-hand side of
Eq.~\ref{eq:epsilon_mv_2} by $r_\Delta \triangleq \left\| \Delta D^{\mathcal{V}_n}v^{\mathcal{V}_n} - R\Delta D^{\mathcal{V}_c}v^{\mathcal{V}_c} \right\|_2. $}

By squaring Eq.~\ref{eq:epsilon_mv_2}, applying $(a+b)^2\leq2a^2+2b^2$, multiplying by the nonnegative weight $w_{ij}(p)$, integrating over $\Omega_{ij}$, and summing over all valid view pairs, we directly obtain the aggregate bound in Eq.~\ref{eq:multi_view_bound}.
Squaring $r_\Delta$ yields
\begin{equation}
    \begin{aligned}
        r_\Delta^2
        &=
        \left(\Delta D^{\mathcal{V}_n}\right)^2
        \left\|v^{\mathcal{V}_n}\right\|_2^2
        +
        \left(\Delta D^{\mathcal{V}_c}\right)^2
        \left\|Rv^{\mathcal{V}_c}\right\|_2^2
        \\
        &\quad
        -
        2\Delta D^{\mathcal{V}_n}
        \Delta D^{\mathcal{V}_c}
        \left(v^{\mathcal{V}_n}\right)^\top
        Rv^{\mathcal{V}_c}.
    \end{aligned}
    \label{eq:delta_u_vn}
\end{equation}
Since $v^{\mathcal{V}_n}$ and $v^{\mathcal{V}_c}$ are unit
camera-ray directions and $R$ is an orthogonal rotation matrix, we
have $\left\|v^{\mathcal{V}_n}\right\|_2 = \left\|Rv^{\mathcal{V}_c}\right\|_2=1$.
Moreover, let $\left(v^{\mathcal{V}_n}\right)^\top Rv^{\mathcal{V}_c}= \cos\theta$, where $\theta$ denotes the angle between $v^{\mathcal{V}_n}$ and
$Rv^{\mathcal{V}_c}$. Equation~\ref{eq:delta_u_vn} then becomes
\begin{equation}
    r_\Delta^2 = \left(\Delta D^{\mathcal{V}_n}\right)^2 + \left(\Delta D^{\mathcal{V}_c}\right)^2 - 2\Delta D^{\mathcal{V}_n} \Delta D^{\mathcal{V}_c} \cos\theta.
    \label{eq:delta_u_vnp}
\end{equation}
Here, $\theta$ is measured in radians. By completing the square,  Eq.~\ref{eq:delta_u_vnp} can be decomposed as
\begin{equation}
    r_\Delta^2 = \left( \Delta D^{\mathcal{V}_n} - \Delta D^{\mathcal{V}_c}\cos\theta \right)^2 + \left(\Delta D^{\mathcal{V}_c}
    \right)^2 \sin^2\theta.
    \label{eq:delta_decomposition}
\end{equation}
From Eqs.~\ref{eq:D_v_n_u} and~\ref{eq:D_v_n_d}, the translation terms cancel exactly when taking the difference between the two predicted depths. Using $\Delta D^{\mathcal{V}_c} = D_u^{\mathcal{V}_c} - 
D_{\mathcal{D}}^{\mathcal{V}_c}$, we obtain
\begin{equation}
    \hat D_u^{\mathcal{V}_n} - \hat D_{\mathcal{D}}^{\mathcal{V}_n} =
    \left(v^{\mathcal{V}_n}\right)^\top R \left(D_u^{\mathcal{V}_c} -     D_{\mathcal{D}}^{\mathcal{V}_c} \right) v^{\mathcal{V}_c} =
    \Delta D^{\mathcal{V}_c}\cos\theta.
    \label{eq:projected_depth_difference}
\end{equation}
Therefore, under the shared-target-ray construction and using Eq.~\ref{eq:projected_depth_difference}, the first term in Eq.~\ref{eq:delta_decomposition} is exactly the squared scalar cross-view depth-difference residual.
Specifically, we define
\begin{equation}
    \begin{aligned}
        \Delta\delta &\triangleq \left| \left(D_u^{\mathcal{V}_n} - 
        D_{\mathcal{D}}^{\mathcal{V}_n} \right) - \left(\hat D_u^{\mathcal{V}_n}
        - \hat D_{\mathcal{D}}^{\mathcal{V}_n} \right) \right|
        \\
        &= \left| \Delta D^{\mathcal{V}_n} - \Delta D^{\mathcal{V}_c}\cos\theta
        \right|.
    \end{aligned}
    \label{eq:delta_residual}
\end{equation}
Accordingly, Eq.~\ref{eq:delta_decomposition} can be equivalently rewritten as
\begin{equation*}
    r_\Delta^2 = \left(\Delta\delta\right)^2 + \left(\Delta D^{\mathcal{V}_c}
    \right)^2 \sin^2\theta.
\end{equation*}
This completes the proof.


\end{proof}
\begin{remark}
Equation~\ref{eq:multi_view_bound} bounds the aggregate multi-view reconstruction error in terms of the rendered-distance-based error and the depth-difference-induced vector residual $r_\Delta$. 
Moreover, Eqs.~\ref{eq:delta_decomposition} and~\ref{eq:delta_residual} give the exact relation
\begin{equation*}
    r_\Delta^2 = \left(\Delta\delta\right)^2 +
    \left(\Delta D^{\mathcal V_c}\right)^2\sin^2\theta,
\end{equation*}
where $\Delta\delta$ denotes the scalar cross-view depth-difference residual.
When the angle-dependent component $\left|\Delta D^{\mathcal V_c}\sin\theta\right|$ is sufficiently small, the exact relationship in the preceding equation can be approximated by
\begin{equation}
    r_\Delta \approx \Delta\delta.
    \label{eq:delta_approx}
\end{equation}
This approximation provides a scalar interpretation of $r_\Delta$ and is not required for the validity of the aggregate bound in Eq.~\ref{eq:multi_view_bound}. Thus, reducing cross-view
depth-difference inconsistency reduces the corresponding scalar residual component and motivates the practical multi-view regularization introduced below.

\end{remark}

\subsubsection{Multi-View Depth-Difference Geometric Constraints}

Motivated by Eqs.~\ref{eq:delta_residual} and~\ref{eq:delta_approx}, we construct a practical cross-view depth consistency regularizer for neighboring views. For $s\in\{u,\mathcal D\}$,
$p_s^{\mathcal V_n}(p)$ denotes the image-plane projection in $\mathcal V_n$ of the point
$D_s^{\mathcal V_c}(p)v^{\mathcal V_c}(p)$ transformed from $\mathcal V_c$.

To determine valid cross-view reprojections, we define the joint validity mask as
\begin{equation}
    \begin{aligned}
        \mathcal M(p)
        \triangleq
        \mathbb I\Big[
        &p_u^{\mathcal V_n}(p)
        \in [0,W-1)\times[0,H-1)
        \\
        &\land\;
        p_{\mathcal D}^{\mathcal V_n}(p)
        \in [0,W-1)\times[0,H-1)
        \\
        &\land\;
        \hat D_u^{\mathcal V_n}(p)>\tau_d
        \land
        \hat D_{\mathcal D}^{\mathcal V_n}(p)>\tau_d
        \Big].
    \end{aligned}
    \label{eq:M}
\end{equation}
Here, $\mathbb I[\cdot]$ denotes the indicator function; $p_u^{\mathcal V_n}(p)$ and
$p_{\mathcal D}^{\mathcal V_n}(p)$ denote the projected coordinates associated with the unbiased depth and rendered distance, respectively; $W$ and $H$ denote the width and height of the neighboring-view depth maps; and $\tau_d$ is the minimum valid-depth threshold.
The valid reprojection domain is
\begin{equation}
    \Omega
    \triangleq
    \left\{
        p\in\Omega_{D_{\mathcal D}}
        \mid
        \mathcal M(p)=1
    \right\}.
    \label{eq:valid_reprojection_domain}
\end{equation}
View pairs with $\Omega=\varnothing$ are excluded from the multi-view
loss computation.

\begin{table*}[b!]
    \centering
    \caption{
    Quantitative reconstruction accuracy on DTU measured by Chamfer distance (mm)$\downarrow$. In the CD columns, red, orange, and yellow indicate the best, second-best, and third-best results, respectively. The superscript $\dagger$ denotes results reproduced by us.}
    \label{tab:quantitativeDTU}
    \resizebox{\textwidth}{!}{
    \begin{tabular}{cccccccccccccccccc}
        \toprule
        \textbf{Method}
        & \textbf{24}
        & \textbf{37}
        & \textbf{40}
        & \textbf{55}
        & \textbf{63}
        & \textbf{65}
        & \textbf{69}
        & \textbf{83}
        & \textbf{97}
        & \textbf{105}
        & \textbf{106}
        & \textbf{110}
        & \textbf{114}
        & \textbf{118}
        & \textbf{122}
        & \textbf{Mean}
        & \textbf{Time} \\
        \midrule

        NeRF~\cite{mildenhall2021nerf}
        & 1.90 & 1.60 & 1.85 & 0.58 & 2.28
        & 1.27 & 1.47 & 1.67 & 2.05 & 1.07
        & 0.88 & 2.53 & 1.06 & 1.15 & 0.96
        & 1.49 & $>$12h \\

        VolSDF~\cite{yariv2021volume}
        & 1.14 & 1.26 & 0.81 & 0.49 & 1.25
        & 0.70 & 0.72 & 1.29 & 1.18 & 0.70
        & 0.66 & 1.08 & 0.42 & 0.61 & 0.55
        & 0.86 & $>$12h \\

        NeuS~\cite{wang2021neus}
        & 1.00 & 1.37 & 0.93 & 0.43 & 1.10
        & 0.65 & 0.57 & 1.48 & 1.09 & 0.83
        & 0.52 & 1.20 & 0.35 & 0.49 & 0.54
        & 0.84 & $>$12h \\

        \midrule

        2DGS~\cite{huang20242d}
        & 0.48
        & 0.82
        & \cellcolor{yellow!50}0.34
        & 0.43
        & 1.35
        & 0.90
        & 0.81
        & 1.27
        & 1.24
        & 0.65
        & 0.70
        & 1.25
        & 0.41
        & 0.66
        & 0.50
        & 0.79
        & 8.5min \\

        GOF~\cite{yu2024gaussian}
        & 0.50
        & 0.83
        & 0.40
        & 0.38
        & 1.35
        & 0.82
        & 0.74
        & 1.26
        & 1.79
        & 0.69
        & 0.71
        & 1.31
        & 0.52
        & 0.64
        & 0.51
        & 0.83
        & 33min \\

        GFSGS~\cite{jiang2025geometry}
        & 0.38
        & 0.65
        & \cellcolor{red!50}0.29
        & 0.36
        & 0.79
        & 0.65
        & 0.64
        & 1.09
        & 1.21
        & \cellcolor{yellow!50}0.59
        & \cellcolor{orange!50}0.47
        & 1.19
        & 0.32
        & 0.42
        & 0.37
        & 0.63
        & 9.9min \\

        GS-Pull~\cite{zhang2024gspull}
        & 0.60
        & 0.66
        & 0.36
        & 0.38
        & \cellcolor{orange!50}0.77
        & 0.74
        & 0.63
        & 1.22
        & \cellcolor{yellow!50}0.96
        & 0.67
        & 0.59
        & 0.58
        & 0.42
        & 0.58
        & 0.53
        & 0.64
        & 22min \\

        GS-UDF~\cite{li2025gaussianudf}
        & 0.62
        & 0.67
        & 0.43
        & 0.42
        & 0.83
        & 0.86
        & 0.72
        & 1.20
        & 1.03
        & 0.68
        & 0.61
        & 0.63
        & 0.43
        & 0.56
        & 0.52
        & 0.68
        & 25min \\

        PGSR~\cite{chen2024pgsr}
        & \cellcolor{yellow!50}0.35
        & \cellcolor{orange!50}0.55
        & 0.40
        & \cellcolor{yellow!50}0.35
        & \cellcolor{yellow!50}0.78
        & \cellcolor{yellow!50}0.58
        & \cellcolor{orange!50}0.49
        & \cellcolor{yellow!50}1.07
        & \cellcolor{red!50}0.64
        & \cellcolor{yellow!50}0.59
        & 0.50
        & \cellcolor{orange!50}0.54
        & 0.33
        & \cellcolor{yellow!50}0.37
        & \cellcolor{orange!50}0.34
        & 0.53
        & 17min \\

        GSDF~\cite{yu2024gsdf}
        & 0.58
        & 0.93
        & 0.46
        & 0.37
        & 1.30
        & 0.77
        & 0.73
        & 1.58
        & 1.28
        & 0.76
        & 0.58
        & 1.22
        & 0.37
        & 0.51
        & 0.51
        & 0.80
        & 32min \\

        QGS~\cite{zhang2025quadratic}
        & 0.42
        & \cellcolor{yellow!50}0.64
        & 0.36
        & 0.38
        & 0.86
        & 0.64
        & \cellcolor{yellow!50}0.50
        & 1.13
        & 0.97
        & 0.61
        & \cellcolor{yellow!50}0.48
        & 0.69
        & 0.34
        & 0.41
        & 0.36
        & 0.59
        & 48min \\

        GSPrior~\cite{noda20263d}
        & \cellcolor{red!50}0.32
        & \cellcolor{orange!50}0.55
        & \cellcolor{orange!50}0.30
        & \cellcolor{red!50}0.31
        & \cellcolor{red!50}0.74
        & \cellcolor{red!50}0.54
        & \cellcolor{red!50}0.48
        & \cellcolor{orange!50}1.06
        & \cellcolor{orange!50}0.69
        & \cellcolor{red!50}0.57
        & \cellcolor{red!50}0.45
        & 0.58
        & \cellcolor{red!50}0.29
        & \cellcolor{red!50}0.35
        & \cellcolor{red!50}0.33
        & \cellcolor{red!50}0.50
        & 42min \\

        GSPrior$^{\dagger}$~\cite{noda20263d}
        & \cellcolor{orange!50}0.33
        & \cellcolor{red!50}0.54
        & 0.37
        & \cellcolor{orange!50}0.34
        & 0.79
        & \cellcolor{yellow!50}0.58
        & \cellcolor{orange!50}0.49
        & \cellcolor{yellow!50}1.07
        & \cellcolor{red!50}0.64
        & \cellcolor{yellow!50}0.59
        & \cellcolor{orange!50}0.47
        & \cellcolor{yellow!50}0.56
        & \cellcolor{yellow!50}0.31
        & 0.38
        & \cellcolor{yellow!50}0.35
        & \cellcolor{yellow!50}0.52
        & 49.9min \\

        Ours
        & \cellcolor{yellow!50}0.35
        & \cellcolor{orange!50}0.55
        & 0.37
        & \cellcolor{orange!50}0.34
        & \cellcolor{orange!50}0.77
        & \cellcolor{orange!50}0.56
        & \cellcolor{orange!50}0.49
        & \cellcolor{red!50}1.04
        & \cellcolor{red!50}0.64
        & \cellcolor{orange!50}0.58
        & \cellcolor{orange!50}0.47
        & \cellcolor{red!50}0.50
        & \cellcolor{orange!50}0.30
        & \cellcolor{orange!50}0.36
        & \cellcolor{red!50}0.33
        & \cellcolor{orange!50}0.51
        & 17.6min \\

        \bottomrule
    \end{tabular}
    }
\end{table*}
\begin{figure*}[t]
    \centering
    \includegraphics[width=0.95\textwidth]{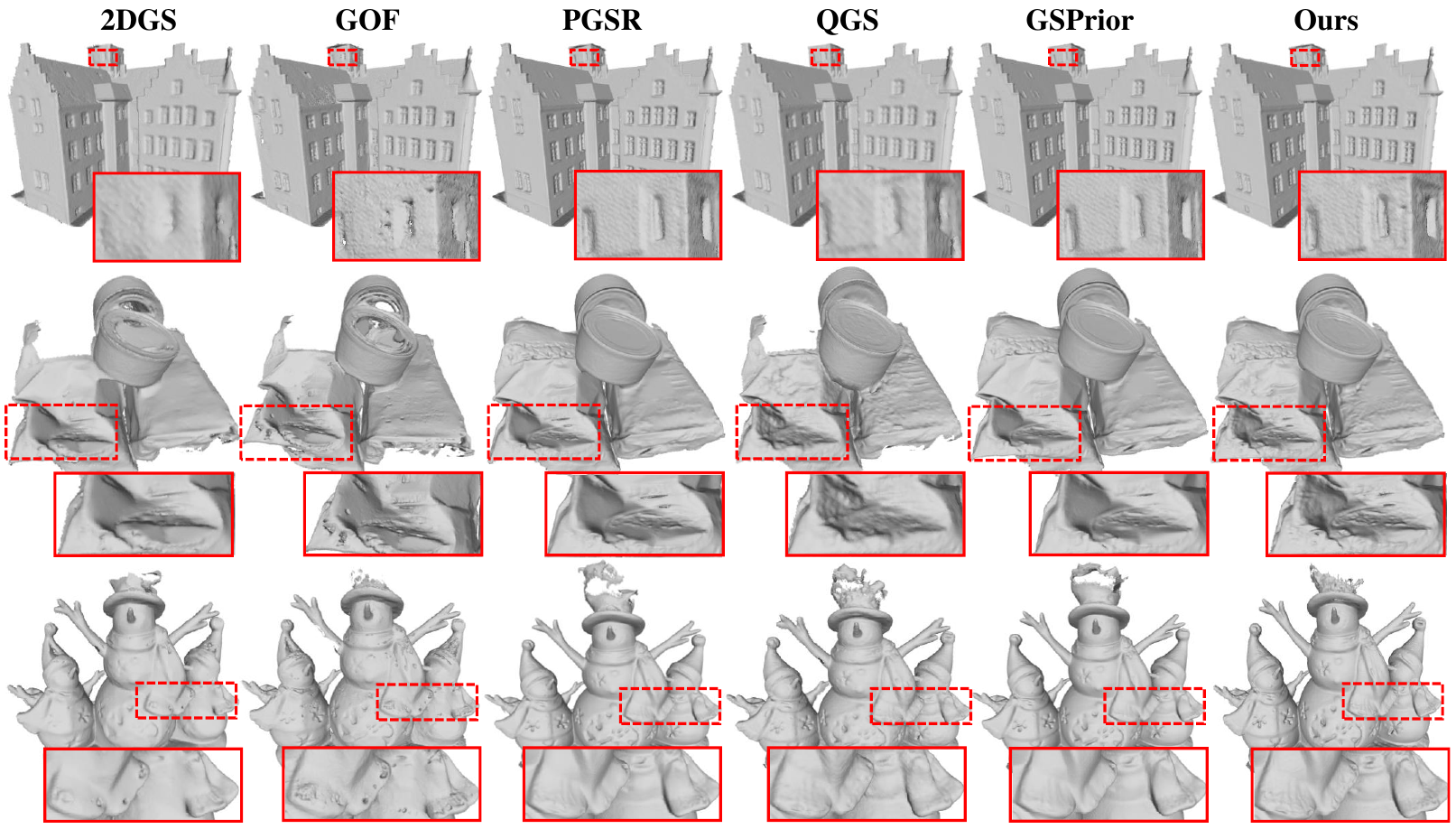}
    \caption{Qualitative surface reconstruction comparison on the DTU dataset. Reconstructed meshes and zoom-in regions are shown to compare local surface details and geometric completeness across different methods.}
    \label{fig:reconstruction}
\end{figure*}
\begin{figure*}[h!]
    \centering
    \includegraphics[width=0.95\textwidth]{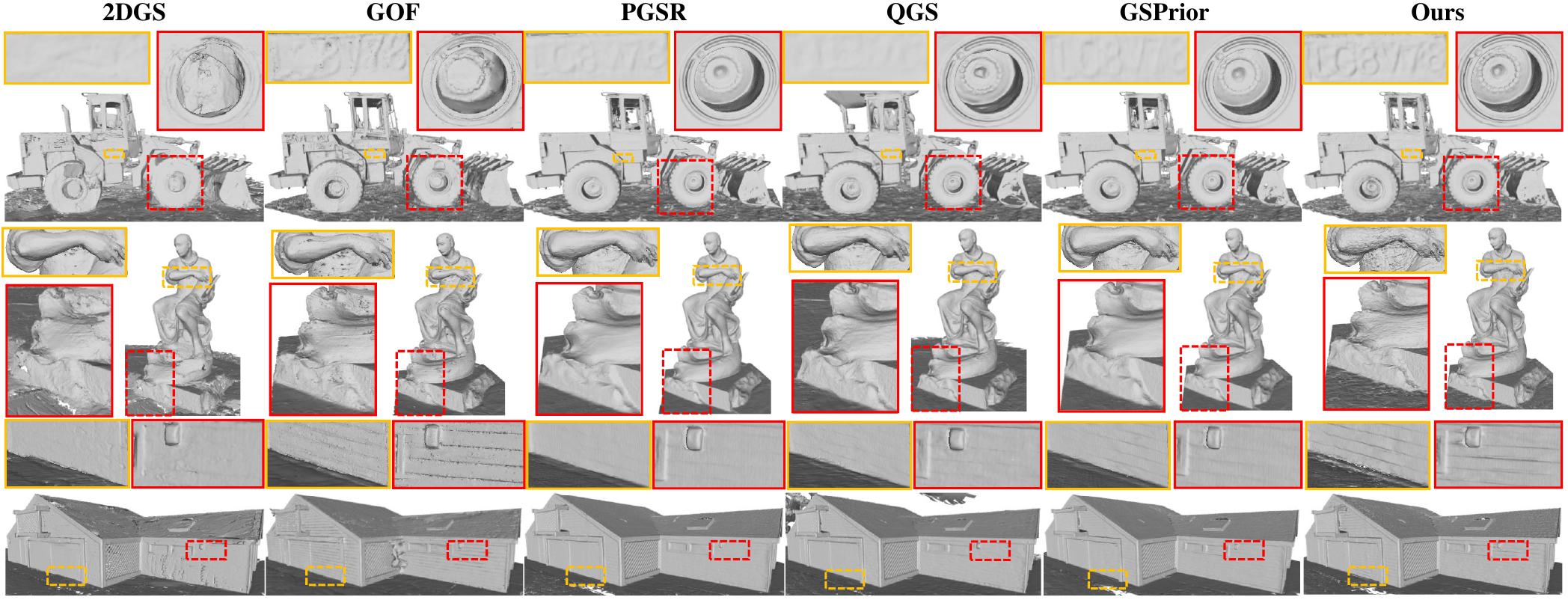}
    \caption{Qualitative surface reconstruction comparison on the Tanks and Temples dataset. Reconstructed meshes and zoom-in regions are shown to compare local surface continuity and geometric details across different methods.}
    \label{fig:tnt_reconstruction}
\end{figure*}

To obtain differentiable depth values at noninteger projected
coordinates, we employ bilinear interpolation:
\begin{equation}
    D_{\pi}(p)
    \triangleq
    \sum_{a=0}^{1}
    \sum_{b=0}^{1}
    \omega_{ab}\,
    D\!\left(
        \lfloor v^\star\rfloor+a,\,
        \lfloor u^\star\rfloor+b
    \right),
    \quad p\in\Omega,
    \label{eq:D_pi}
\end{equation}
where $(u^\star,v^\star)$ denotes a projected pixel coordinate,
$\Delta u=u^\star-\lfloor u^\star\rfloor$,
$\Delta v=v^\star-\lfloor v^\star\rfloor$, and
    $\omega_{ab}
    =
    \big[(1-a)(1-\Delta v)+a\Delta v\big]
    \big[(1-b)(1-\Delta u)+b\Delta u\big]$.
Applying Eq.~\ref{eq:D_pi} to $D_u^{\mathcal V_n}$ at
$p_u^{\mathcal V_n}(p)$ and to
$D_{\mathcal D}^{\mathcal V_n}$ at
$p_{\mathcal D}^{\mathcal V_n}(p)$ yields
$D_{u,\pi}^{\mathcal V_n}(p)$ and
$D_{\mathcal D,\pi}^{\mathcal V_n}(p)$, respectively.

The two cross-view depth discrepancies are defined as
\begin{align}
    \delta_u(p)
    &\triangleq
    \left|
        \hat D_u^{\mathcal V_n}(p)
        -
        D_{u,\pi}^{\mathcal V_n}(p)
    \right|,
    \label{eq:delta_u}
    \\
    \delta_{\mathcal D}(p)
    &\triangleq
    \left|
        \hat D_{\mathcal D}^{\mathcal V_n}(p)
        -
        D_{\mathcal D,\pi}^{\mathcal V_n}(p)
    \right|.
    \label{eq:delta_D}
\end{align}
Eqs.~\ref{eq:delta_u} and~\ref{eq:delta_D} measure the discrepancies between the current-view depth predictions transformed into the neighboring view and the corresponding depths sampled from the neighboring-view depth maps.

To connect these practical discrepancies with the scalar residual in
Eq.~\ref{eq:delta_residual}, we define its sampled counterpart as
\begin{equation}
    \begin{aligned}
        \Delta\delta_{\pi}(p)
        \triangleq
        \Bigl|
        \left(D_{u,\pi}^{\mathcal V_n}(p)
        -
        D_{\mathcal D,\pi}^{\mathcal V_n}(p) \right)
        -
        \left( \hat D_u^{\mathcal V_n}(p)
        -
        \hat D_{\mathcal D}^{\mathcal V_n}(p)\right)
        \Bigr|.
    \end{aligned}
    \label{eq:sampled_delta}
\end{equation}
Applying the triangle inequality to Eq.~\ref{eq:sampled_delta} and using Eqs.~\ref{eq:delta_u} and~\ref{eq:delta_D} yield
\begin{equation}
    \Delta\delta_{\pi}(p)
    \leq
    \delta_u(p)+\delta_{\mathcal D}(p).
    \label{eq:sampled_delta_triangle}
\end{equation}
Moreover, for any $\alpha\in(0,1)$,
\begin{equation}
    \delta_u(p)+\delta_{\mathcal D}(p)
    \leq
    \frac{
        \alpha\delta_u(p)
        +(1-\alpha)\delta_{\mathcal D}(p)
    }{
        \min\{\alpha,1-\alpha\}
    }.
    \label{eq:weighted_delta_bound}
\end{equation}
Eqs.~\ref{eq:sampled_delta_triangle}
and~\ref{eq:weighted_delta_bound} establish that a positive weighted combination of the two depth discrepancies controls $\Delta\delta_{\pi}(p)$ up to the constant factor $1/\min\{\alpha,1-\alpha\}$.

Accordingly, we define the cross-view depth consistency regularizer as
\begin{equation}
    \mathcal L_{mv\_D}
    =
    \frac{1}{|\Omega|}
    \sum_{p\in\Omega}
    \left[
        \alpha\delta_u(p)
        +
        (1-\alpha)\delta_{\mathcal D}(p)
    \right],
    \label{eq:loss_mv_d}
\end{equation}
where $\alpha$ balances the two depth discrepancies and is set to $\alpha=0.8$. Since $\alpha\in(0,1)$, both terms are assigned positive weights. Thus, minimizing $\mathcal L_{mv\_D}$ jointly penalizes the
two reprojection discrepancies and serves as a practical surrogate for reducing the sampled cross-view depth-difference residual $\Delta\delta_{\pi}(p)$. 

\subsection{Optimization}

The proposed regularizers are jointly optimized with the image reconstruction loss $\mathcal L_{rgb}$, flattened-Gaussian loss $\mathcal L_s$, and multi-view photometric consistency loss $\mathcal L_{mvgeom}$:
\begin{equation}
    \mathcal L
    =
    \lambda_1\mathcal L_{rgb}
    +\lambda_2\mathcal L_s
    +\lambda_3\mathcal L_{mvgeom}
    +\lambda_4\mathcal L_{svgeo}
    +\lambda_5\mathcal L_{mv\_D}.
\end{equation}
Here, $\lambda_1$--$\lambda_5$ denote the corresponding loss weights; $\lambda_1$--$\lambda_4$ retain the PGSR settings, while $\lambda_5$ controls the proposed cross-view regularizer.

\section{EXPERIMENTS}
\subsection{Datasets and Implementation Details}

We conduct experiments on three public benchmarks: Mip-NeRF360~\cite{barron2022mip}, Tanks and
Temples~\cite{knapitsch2017tanks}, and DTU~\cite{jensen2014large}.
Novel view synthesis is evaluated on Mip-NeRF360, while surface reconstruction is evaluated on DTU and Tanks and Temples.
Rendering quality is measured using peak signal-to-noise ratio (PSNR), structural similarity (SSIM), and learned perceptual image patch similarity (LPIPS). 
Surface reconstruction is evaluated using Chamfer distance (CD) on DTU and the benchmark F1-score on Tanks and Temples.

All experiments are conducted on a single NVIDIA GeForce RTX 4090 GPU. Unless otherwise specified, the training strategy and hyperparameters follow those of PGSR~\cite{chen2024pgsr}, and each scene is optimized for 30,000 iterations. Following PGSR, the minimum valid-depth threshold is set to $\tau_d=0.1$, and the loss weights $\lambda_1$, $\lambda_2$, $\lambda_3$, and $\lambda_4$ adopt the baseline settings. For the proposed components, the numerical stabilization constant in Eq.~\ref{eq:delta_W(p)} is set to $\sigma=10^{-6}$, and the cross-view depth consistency regularization weight is set to $\lambda_5=0.008$. After optimization, depth maps rendered from the optimized Gaussian representation are fused into a truncated signed distance function (TSDF) volume, from which a surface mesh is extracted for reconstruction evaluation.

\begin{figure*}[b]
    \centering
    \includegraphics[width=\textwidth]{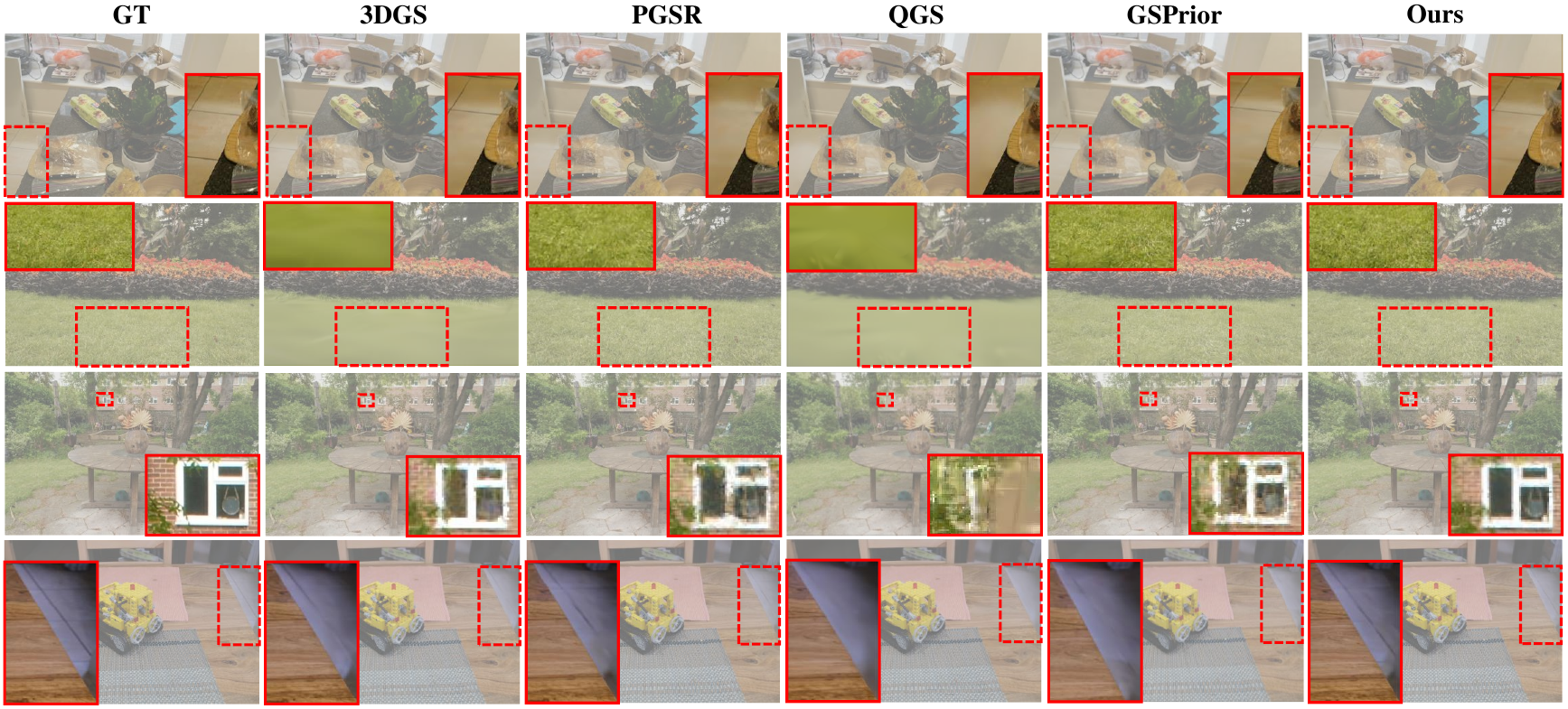}  
    \caption{Qualitative comparison on the Mip-NeRF360 dataset. Rendered images from different methods are shown with enlarged local regions marked by red boxes. The results suggest that our method preserves fine image details and local structures in the selected scenes.}
    \label{fig:render}
\end{figure*}
\begin{table}[t]
\centering
\caption{Quantitative F1-score comparison on Tanks and Temples. Bold indicates the best result within each class.}
\label{tab:tnt_fscore}
\tiny
\setlength{\tabcolsep}{1.0pt}
\renewcommand{\arraystretch}{0.88}
\makebox[\columnwidth][c]{%
\resizebox{0.9\columnwidth}{!}{%
\begin{tabular}{@{}c|c|cccccc|c@{}}
\toprule
Class & Method & Barn & Caterpillar & Courthouse & Ignatius & Meetingroom & Truck & Mean \\
\midrule
\multirow{3}{*}{\rotatebox[origin=c]{90}{Implicit}}
& NeuS~\cite{wang2021neus} 
& 0.29 & 0.29 & 0.17 & 0.83 & 0.24 & 0.45 & 0.378 \\
& Neuralangelo~\cite{li2023neuralangelo} 
& \textbf{0.70} & \textbf{0.36} & \textbf{0.28} & \textbf{0.89} & \textbf{0.32} & \textbf{0.48} & \textbf{0.505} \\
& MonoSDF~\cite{yu2022monosdf} 
& 0.49 & 0.31 & 0.12 & 0.78 & 0.23 & 0.42 & 0.392 \\
\midrule
\multirow{8}{*}{\rotatebox[origin=c]{90}{Explicit}}
& 3DGS~\cite{kerbl20233d} 
& 0.13 & 0.08 & 0.09 & 0.04 & 0.01 & 0.19 & 0.090 \\
& 2DGS~\cite{huang20242d} 
& 0.36 & 0.23 & 0.13 & 0.44 & 0.16 & 0.26 & 0.263 \\
& GOF~\cite{yu2024gaussian} 
& 0.51 & 0.41 & \textbf{0.28} & 0.68 & 0.28 & 0.59 & 0.458 \\
& PGSR~\cite{chen2024pgsr} 
& \textbf{0.66} & 0.41 & 0.21 & 0.80 & 0.29 & 0.60 & 0.495 \\
& GS-Pull~\cite{zhang2024gspull} 
& 0.60 & 0.37 & 0.16 & 0.71 & 0.22 & 0.52 & 0.430 \\
& QGS~\cite{zhang2025quadratic} 
& 0.55 & 0.40 & \textbf{0.28} & \textbf{0.81} & \textbf{0.31} & \textbf{0.64} & 0.500 \\
& GSPrior~\cite{noda20263d} 
& \textbf{0.66} & \textbf{0.43} & 0.20 & 0.80 & \textbf{0.31} & \textbf{0.64} & \textbf{0.507} \\
& Ours 
& 0.64 & \textbf{0.43} & 0.21 & \textbf{0.81} & 0.30 & 0.63 & 0.503 \\
\bottomrule
\end{tabular}%
}%
}
\end{table}

\begin{table*}[t]
    \centering
    \caption{
    Novel-view-synthesis results on Mip-NeRF360. Red, orange, and yellow indicate the best, second-best, and third-best results, respectively; $\dagger$ denotes reproduced
    results.}
    \label{tab:renderComparison}
    \resizebox{\textwidth}{!}{%
    \begin{tabular}{c*{9}{c}}
        \toprule
        \multirow{2}{*}{Method}
        & \multicolumn{3}{c}{Outdoor scenes}
        & \multicolumn{3}{c}{Indoor scenes}
        & \multicolumn{3}{c}{Overall average} \\
        \cmidrule(lr){2-4}
        \cmidrule(lr){5-7}
        \cmidrule(lr){8-10}
        & SSIM$\uparrow$
        & PSNR$\uparrow$
        & LPIPS$\downarrow$
        & SSIM$\uparrow$
        & PSNR$\uparrow$
        & LPIPS$\downarrow$
        & SSIM$\uparrow$
        & PSNR$\uparrow$
        & LPIPS$\downarrow$ \\
        \midrule
        3DGS~\cite{kerbl20233d} 
        & 0.742 
        & \cellcolor{red!50}25.07 
        & 0.232 
        & \cellcolor{orange!50}0.931 
        & \cellcolor{red!50}31.18 
        & 0.164 
        & 0.837 
        & \cellcolor{red!50}28.13 
        & 0.198 \\

        2DGS~\cite{huang20242d} 
        & 0.706 
        & 24.21 
        & 0.284 
        & 0.912 
        & 30.15 
        & 0.212 
        & 0.809 
        & 27.18 
        & 0.248 \\

        GOF~\cite{yu2024gaussian} 
        & 0.746 
        & \cellcolor{orange!50}24.81 
        & 0.208 
        & 0.917 
        & 30.40 
        & 0.189 
        & 0.832 
        & 27.61 
        & 0.199 \\

        GFSGS ~\cite{jiang2025geometry}
        & 0.734 
        & 24.40 
        & 0.224 
        & 0.916 
        & 29.93 
        & 0.194 
        & 0.825 
        & 27.17 
        & 0.209 \\

        GS-UDF~\cite{li2025gaussianudf} 
        & 0.698 
        & 23.84 
        & 0.292 
        & 0.906 
        & 29.16 
        & 0.219 
        & 0.802 
        & 26.50 
        & 0.255 \\

        PGSR~\cite{chen2024pgsr} 
        & \cellcolor{yellow!50}0.752 
        & 24.74 
        & 0.203 
        & \cellcolor{yellow!50}0.929 
        & 30.16 
        & \cellcolor{yellow!50}0.159 
        & \cellcolor{yellow!50}0.841 
        & 27.45 
        & 0.181 \\

        QGS~\cite{zhang2025quadratic} 
        & 0.706 
        & 23.32 
        & 0.242 
        & 0.919 
        & 30.45 
        & 0.184 
        & 0.813 
        & 26.89 
        & 0.213 \\

        GSPrior~\cite{noda20263d} 
        & \cellcolor{orange!50}0.754 
        & \cellcolor{orange!50}24.81 
        & \cellcolor{orange!50}0.200 
        & \cellcolor{red!50}0.933 
        & \cellcolor{yellow!50}30.56 
        & \cellcolor{orange!50}0.155 
        & \cellcolor{red!50}0.844 
        & \cellcolor{yellow!50}27.69 
        & \cellcolor{orange!50}0.178 \\

        GSPrior$^{\dagger}$~\cite{noda20263d}  
        & \cellcolor{red!50}0.755 
        & \cellcolor{yellow!50}24.77 
        & \cellcolor{yellow!50}0.201 
        & \cellcolor{orange!50}0.931 
        & 30.39 
        & \cellcolor{yellow!50}0.159 
        & \cellcolor{orange!50}0.843 
        & 27.58 
        & \cellcolor{yellow!50}0.180 \\

        Ours 
        & 0.749 
        & 24.75 
        & \cellcolor{red!50}0.199 
        & \cellcolor{red!50}0.933 
        & \cellcolor{orange!50}30.68 
        & \cellcolor{red!50}0.151 
        & \cellcolor{yellow!50}0.841 
        & \cellcolor{orange!50}27.72 
        & \cellcolor{red!50}0.175 \\
        \bottomrule
    \end{tabular}%
    }
\end{table*}
\begin{figure*}[h!]
    \centering
    \includegraphics[width=0.80\textwidth]
    {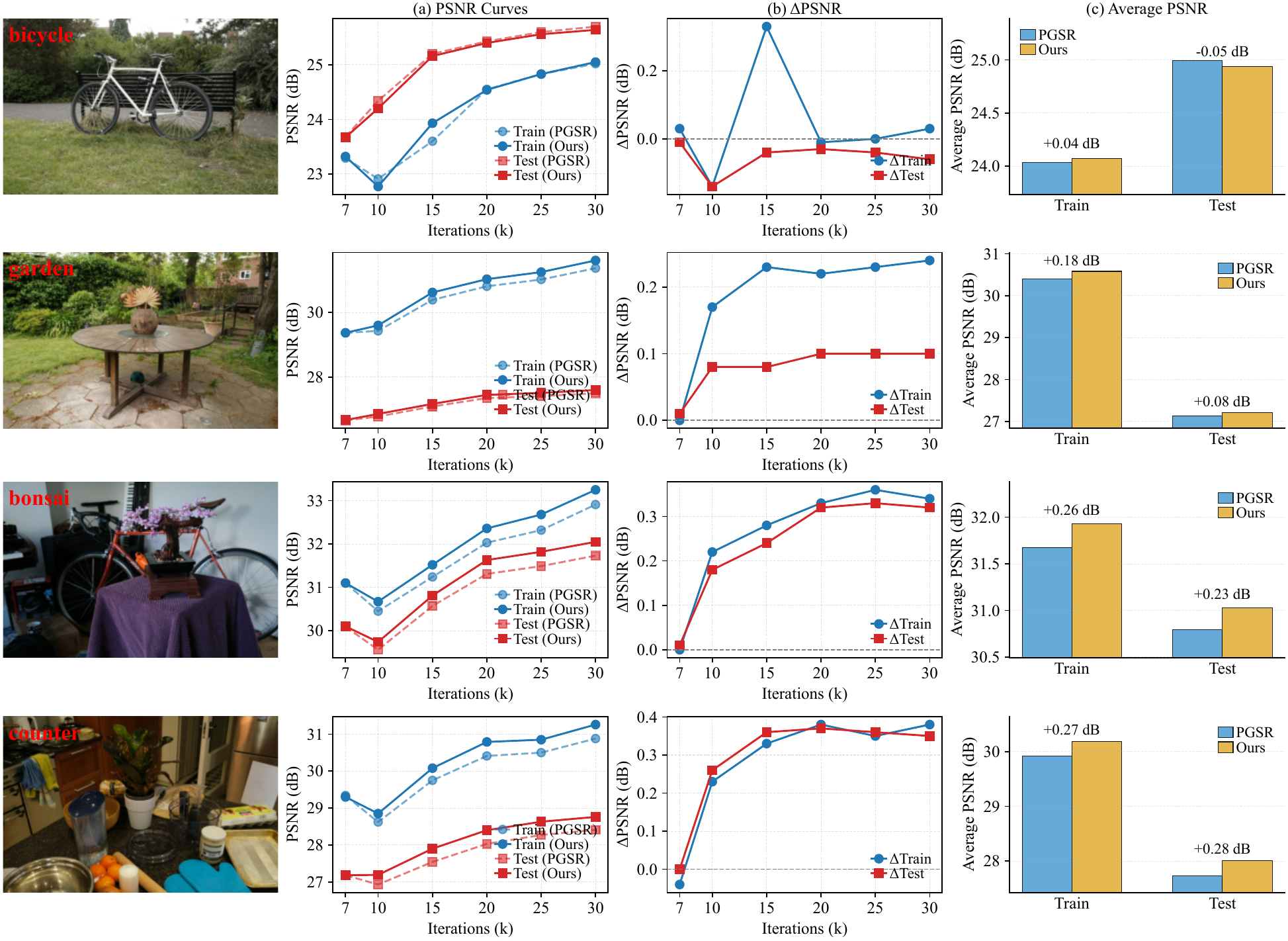}
    \caption{
    Optimization dynamics compared with PGSR.
    (a) Training- and test-view PSNR curves;
    (b) per-iteration $\Delta\mathrm{PSNR}$;
    (c) average training- and test-view PSNR.}
    \label{fig:dynamic_compare}
\end{figure*}

\subsection{Surface Reconstruction Comparisons}

\noindent \textbf{Comparisons on DTU.} We evaluate the surface reconstruction performance of the proposed method on the DTU dataset~\cite{jensen2014large} and compare it with representative neural implicit methods and recent Gaussian-splatting-based reconstruction approaches, as listed in Table~\ref{tab:quantitativeDTU}. 
The quantitative results show that our method obtains a mean Chamfer distance of 0.51 mm, ranking second in terms of average reconstruction accuracy among the compared methods. 
Compared with PGSR, our method reduces the mean Chamfer distance from 0.53 mm to 0.51 mm under a similar experimental setting. 
Overall, these results indicate that our method achieves competitive reconstruction accuracy under the reported experimental setting.

Fig.~\ref{fig:reconstruction} presents qualitative comparisons of reconstructed meshes with enlarged local regions. The visual results suggest that our method can maintain relatively continuous surfaces and preserve local geometric details in challenging regions, which is consistent with the quantitative comparison.

\noindent \textbf{Comparisons on Tanks and Temples.}
We further evaluate our method on the large-scale Tanks and Temples dataset to assess its reconstruction generalization beyond the controlled DTU setting. 
As shown in Table~\ref{tab:tnt_fscore}, our method achieves a competitive mean F1-score and shows comparable performance to recent Gaussian-splatting-based reconstruction methods. 
Fig.~\ref{fig:tnt_reconstruction} presents qualitative comparisons of reconstructed meshes and enlarged local regions. The results suggest that our method can produce relatively continuous surfaces and preserve local geometric details in challenging large-scale scenes.

\subsection{Novel View Synthesis Comparisons}
Table~\ref{tab:renderComparison} and Fig.~\ref{fig:render} present the quantitative and qualitative novel view synthesis results on the Mip-NeRF360 dataset. We evaluate rendering quality using PSNR, SSIM, and LPIPS, and compare our method with representative and recent Gaussian-splatting-based approaches. 

As shown in Table~\ref{tab:renderComparison}, our method achieves the lowest LPIPS across outdoor scenes, indoor scenes, and the overall average, indicating favorable perceptual rendering quality. It also obtains the second-best overall PSNR and competitive SSIM compared with recent methods. These results suggest that the proposed geometric regularization maintains competitive novel view synthesis quality while improving perceptual rendering metrics.
Fig.~\ref{fig:render} further provides qualitative comparisons with enlarged local regions, where our method preserves fine image details and local structures in the selected scenes. 

\subsection{Optimization Dynamics}
We record the training- and test-view PSNR at 7k, 10k, 15k, 20k, 25k, and 30k iterations and compare DepthGS with PGSR in Fig.~\ref{fig:dynamic_compare}. The first column shows the PSNR evolution, the second reports $\Delta\mathrm{PSNR}
=\mathrm{PSNR}_{\mathrm{Ours}} -\mathrm{PSNR}_{\mathrm{PGSR}}$, and the third compares the average training- and test-view PSNR. 
DepthGS generally achieves higher PSNR at most recorded iterations, with faster improvement during 7k--15k iterations in several scenes. 
The maximum training/test gains reach 0.38/0.37~dB in counter and 0.36/0.33~dB in bonsai.

The average training/test gains are 0.27/0.28~dB in counter, 0.26/0.23~dB in bonsai, and 0.18/0.08~dB in garden. 
In the more challenging bicycle scene, the training PSNR increases by 0.04~dB, whereas the test PSNR decreases by 0.05~dB.
This degradation may result from complex illumination, wide viewpoint variations, and fragmented geometry, which reduce the reliability of normal and depth estimates. Overall, the results indicate favorable but scene-dependent optimization behavior rather than universally improved convergence.

\begin{figure*}[t]
    \centering
    \includegraphics[width=0.80\textwidth]{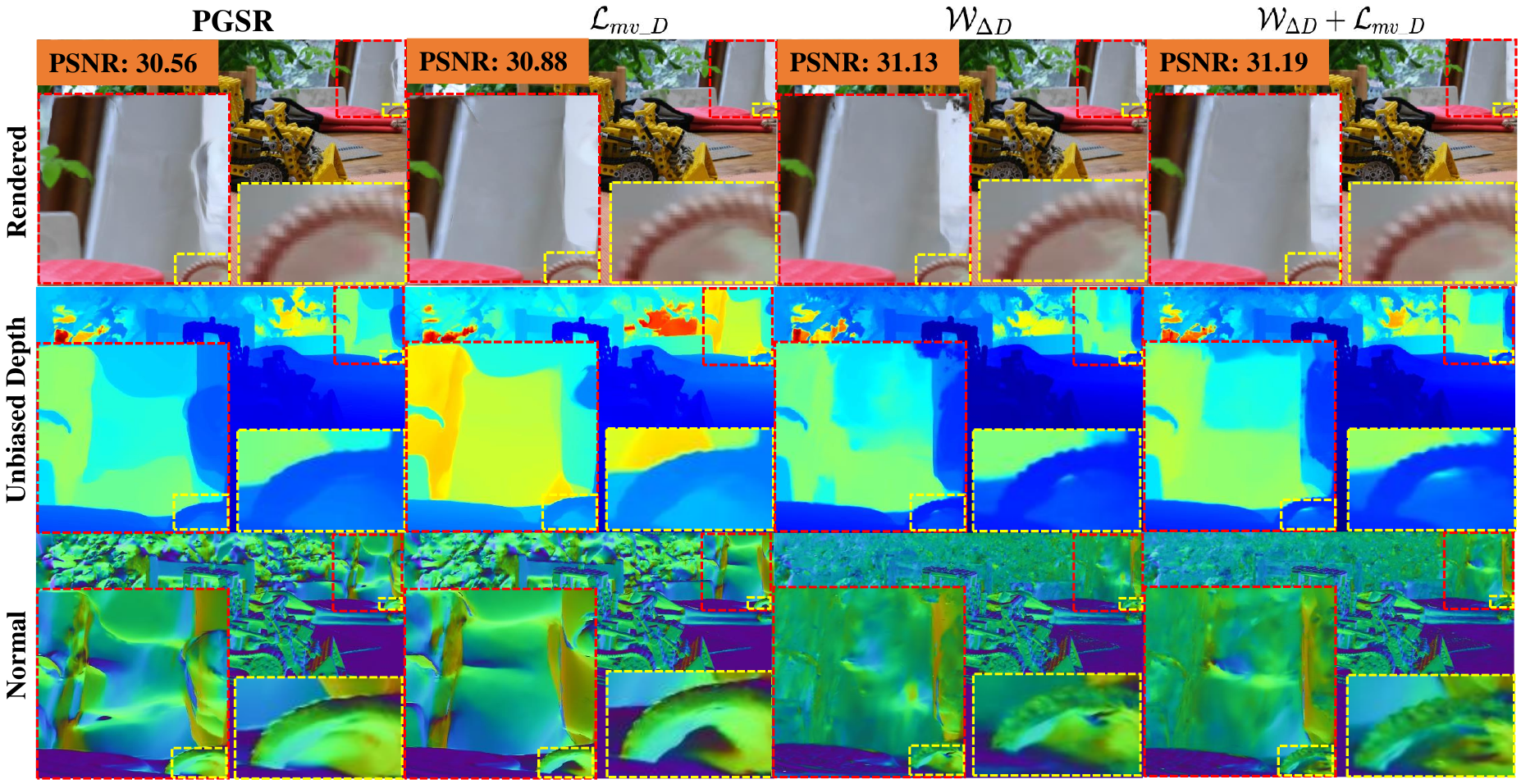}  
    \caption{Novel-view-synthesis ablation results. The rows show RGB, unbiased depth, and normal maps; the columns show PGSR, the two individual components, and the full model.}
    \label{fig:ablation_render}
\end{figure*}

\subsection{Ablation Studies}

\noindent
\textbf{Ablation settings and results.}
We evaluate the depth-difference-guided weight $\mathcal W_{\Delta D}$ and the cross-view depth consistency regularizer $\mathcal L_{mv\_D}$. The image-gradient factor $\overline{\nabla I}^{\,2}$ is kept unchanged whenever the single-view normal constraint is used, so the ablation isolates the contribution of $\mathcal W_{\Delta D}$. In Tables~\ref{tab:nerf_ablation} and~\ref{tab:dtu_ablation},
``$\times$'' and ``$\checkmark$'' indicate the exclusion and inclusion of each component, respectively.

Table~\ref{tab:nerf_ablation} shows that $\mathcal W_{\Delta D}$ primarily improves image-level rendering metrics, while the full model maintains comparable rendering quality.
Table~\ref{tab:dtu_ablation} shows that the proposed components reduce CD from 0.53 to 0.51~mm, with average and peak memory usages of 6.77 and 7.50~GB for the full model. 
The complementary effect of $\mathcal L_{mv\_D}$ is further evaluated using $E_{\mathrm{MVC}}$ below. 

\begin{table}[t]
\centering
\caption{Ablation study on the Mip-NeRF360 dataset. Best results are in bold.}
\label{tab:nerf_ablation}
\renewcommand{\arraystretch}{1.15}
\resizebox{0.85\columnwidth}{!}{
\begin{tabular}{ccccc}
\toprule
$\mathcal{W}_{\Delta D}$ 
& $\mathcal{L}_{mv\_D}$ 
& \textbf{PSNR}$\uparrow$ 
& \textbf{SSIM}$\uparrow$ 
& \textbf{LPIPS}$\downarrow$ \\
\midrule
$\times$     & $\times$     & 27.45          & 0.841          & 0.181 \\
$\checkmark$ & $\times$     & \textbf{27.75} & \textbf{0.842} & \textbf{0.175} \\
$\times$     & $\checkmark$ & 27.51          & \textbf{0.842} & 0.180 \\
$\checkmark$ & $\checkmark$ & 27.73          & 0.841          & \textbf{0.175} \\
\bottomrule
\end{tabular}}
\end{table}

\noindent\textbf{Qualitative ablation analysis.}
Fig.~\ref{fig:ablation_render} compares the rendered RGB images, unbiased depth maps, and normal maps produced by PGSR and the three ablation variants. Introducing either $\mathcal L_{mv\_D}$ or $\mathcal W_{\Delta D}$ improves local boundary preservation and geometric coherence in the highlighted regions. 
Their combination achieves the highest PSNR of 31.19~dB for the illustrated view and
provides a favorable overall result across the three representations.
\begin{table}[h]
\centering
\caption{Ablation study on the DTU dataset (CD: mm; memory: GB).}
\label{tab:dtu_ablation}
\renewcommand{\arraystretch}{1.18}
\resizebox{\columnwidth}{!}{
\begin{tabular}{cccccc}
    \toprule
    \textbf{Method} 
    & $\mathcal{W}_{\Delta D}$ 
    & $\mathcal{L}_{mv\_D}$ 
    & \textbf{CD}$\downarrow$ 
    & \textbf{Avg. Mem.}$\downarrow$ 
    & \textbf{Peak Mem.}$\downarrow$ \\
    \midrule
    PGSR & $\times$ & $\times$ & 0.53 & 6.41 & 7.12 \\
    + $\mathcal{L}_{mv\_D}$ & $\times$ & $\checkmark$ & 0.52 & 6.89 & 7.98 \\
    + $\mathcal{W}_{\Delta D}$ & $\checkmark$ & $\times$ & \textbf{0.51} & 6.74 
    & 7.43 \\
    Ours & $\checkmark$ & $\checkmark$ & \textbf{0.51} & 6.77 & 7.50 \\
    \bottomrule
\end{tabular}}
\end{table}
\begin{figure}[t]
    \centering
    \includegraphics[width=0.9\linewidth]{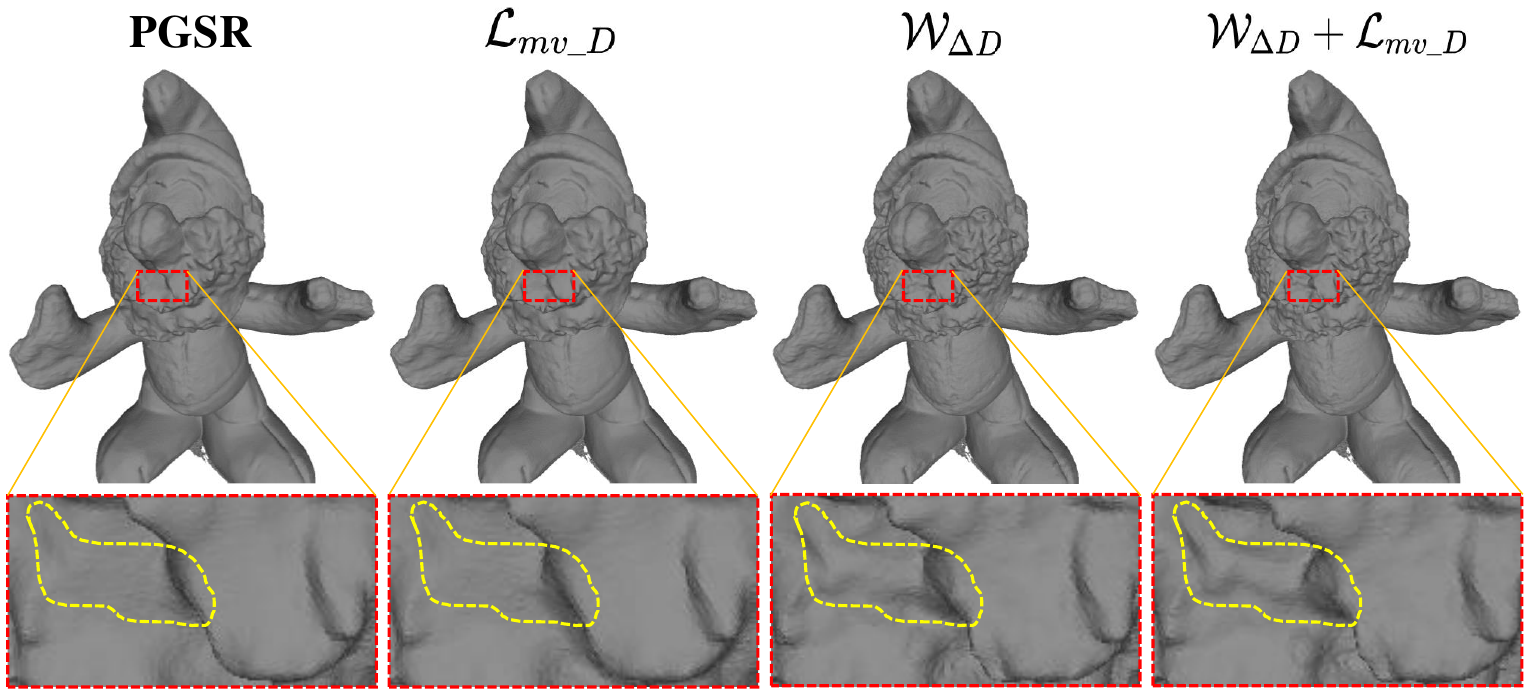}
    \caption{3D reconstruction ablation study. The red dashed box indicates the magnified region. The comparison of local structures within the highlighted region illustrates the effects of different components on local contour preservation and geometric details.}
    \label{fig:ablation_reconstruction}
\end{figure}

Fig.~\ref{fig:ablation_reconstruction} shows that the proposed components generally produce clearer local contours and better-defined geometric structures. These results are consistent with their intended roles: $\mathcal W_{\Delta D}$ adapts the local normal constraint according to geometric reliability, while $\mathcal L_{mv\_D}$ reduces cross-view depth discrepancies.

\noindent
\textbf{Analysis of cross-view depth consistency regularization.}
Using the sampled cross-view depth-difference residual defined in Eq.~\ref{eq:sampled_delta}, we define the multi-view consistency error for each valid view pair as
\begin{equation}
    \begin{aligned}
        E_{\mathrm{MVC}} =\frac{1}{|\Omega|} \sum_{p\in\Omega}  \Big|  &  \left(  D_{u,\pi}^{\mathcal V_n}(p)  - D_{\mathcal D,\pi}^{\mathcal V_n}(p)  \right)
        \\
        &-
        \left( \hat D_u^{\mathcal V_n}(p)  - \hat D_{\mathcal D}^{\mathcal V_n}(p)
        \right)  \Big|.
    \end{aligned}
\end{equation}
A lower $E_{\mathrm{MVC}}$ indicates better cross-view depth-difference consistency. View pairs with $\Omega=\varnothing$ are excluded from the evaluation.
\begin{figure}[h]
    \centering
    \includegraphics[width=\linewidth]{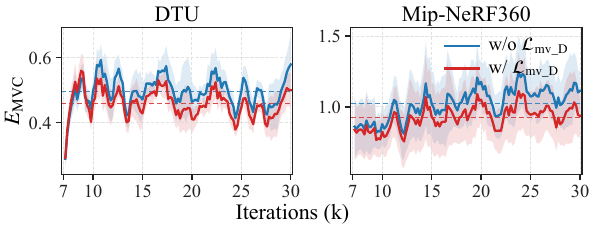}
    \caption{Mean $E_{\mathrm{MVC}}$ on DTU and Mip-NeRF360. Shaded regions denote standard errors across scenes.}
    \label{fig:ablation_emvc_curve}
\end{figure}

As shown in Fig.~\ref{fig:ablation_emvc_curve}, the averaged $E_{\mathrm{MVC}}$ curves on DTU and Mip-NeRF360 are generally lower with $\mathcal{L}_{mv\_D}$ than without it, indicating that the proposed cross-view depth consistency regularization helps reduce cross-view depth-difference inconsistency. Since $\mathcal{L}_{mv\_D}$ mainly targets cross-view geometric consistency, $E_{\mathrm{MVC}}$ provides a direct diagnostic measure for evaluating its intended effect.


\section{CONCLUSION}
In this paper, we have proposed DepthGS, a depth-guided Gaussian splatting framework with adaptive regularization. More specifically, we have derived theoretical relations between depth information and reconstruction error in both single-view and multi-view settings and established corresponding upper bounds on the reconstruction error. We have further proposed depth-driven optimization frameworks for single-view and multi-view settings that use depth-derived geometric cues to regularize the optimization and improve geometric consistency. Experimental results on multiple public datasets, including DTU, Mip-NeRF360, and Tanks and Temples, demonstrate competitive reconstruction and rendering performance in both quantitative metrics and qualitative comparisons. Future research will focus on addressing challenges in complex outdoor environments, especially under strong illumination variations, noisy geometry, and unstructured scene layouts, to further improve the robustness of the method in real-world applications.


\bibliographystyle{IEEEtran} 
\bibliography{IEEEfull}

\makeatletter

\def\@IEEEBIOskipN{1.5\baselineskip}

\patchcmd{\IEEEbiography}
  {\vskip \@IEEEBIOskipN plus 1fil minus 0\baselineskip}
  {\vskip \@IEEEBIOskipN}
  {}
  {\PackageWarning{IEEEbiography}
   {Failed to patch the biography spacing}}

\makeatother

\end{document}